\documentclass[preprint,12pt]{elsarticle}

\usepackage{graphicx}
\usepackage{hyperref}
\usepackage{amssymb}
\usepackage{longtable}
\usepackage{multirow}
\usepackage{appendix}
\usepackage{array}
\newcolumntype{C}[1]{>{\centering\arraybackslash}m{#1}}
\usepackage[T1]{fontenc}
\usepackage{lscape}
\usepackage{lineno}
\usepackage[graphicx]{realboxes}
\usepackage{rotating}
\usepackage{afterpage}
\usepackage{chngpage}
\usepackage{longtable}
\usepackage{caption}
\usepackage{booktabs}
\usepackage{geometry}
\usepackage{tabularx}
\usepackage{rotating}
\usepackage{amsmath}
\usepackage{bbding}
\usepackage{afterpage}
\usepackage{todonotes}

\providecommand{\tabularnewline}{\\}
\makeatother

\usepackage[usestackEOL]{stackengine}
\journal{Journal Name}

\begin{document}

\begin{frontmatter}

\title{From Concept Drift to Model Degradation: An Overview on Performance-Aware Drift Detectors}

\author{Firas Bayram}
\address{Department of Mathematics and Computer Science, Karlstad University\\ 651 88 Karlstad, Sweden\\ email: firas.bayram@kau.se}
\author{Bestoun S. Ahmed}
\address{Department of Mathematics and Computer Science, Karlstad University\\ 651 88 Karlstad, Sweden\\ email: bestoun@kau.se}
\author{Andreas Kassler}
\address{Department of Mathematics and Computer Science, Karlstad University\\ 651 88 Karlstad, Sweden\\ email: andreas.kassler@kau.se}

\begin{abstract}
The dynamicity of real-world systems poses a significant challenge to deployed predictive machine learning (ML) models. Changes in the system on which the ML model has been trained may lead to performance degradation during the system's life cycle. Recent advances that study non-stationary environments have mainly focused on identifying and addressing such changes caused by a phenomenon called concept drift. Different terms have been used in the literature to refer to the same type of concept drift and the same term for various types. This lack of unified terminology is set out to create confusion on distinguishing between different concept drift variants. In this paper, we start by grouping concept drift types by their mathematical definitions and survey the different terms used in the literature to build a consolidated taxonomy of the field. We also review and classify performance-based concept drift detection methods proposed in the last decade. These methods utilize the predictive model's performance degradation to signal substantial changes in the systems. The classification is outlined in a hierarchical diagram to provide an orderly navigation between the methods. We present a comprehensive analysis of the main attributes and strategies for tracking and evaluating the model's performance in the predictive system. The paper concludes by discussing open research challenges and possible research directions.

\end{abstract}


\end{frontmatter}

\begingroup
\tiny
\renewcommand{\arraystretch}{1}
\begin{longtable}{l|l}
\caption{Acronyms used in the paper}
\tabularnewline
\endfirsthead
\caption{Acronyms used in the paper}
\tabularnewline
\endhead
\hline 
\textbf{Acronym} & \textbf{Referring to/ Definition}\tabularnewline
\hline 
AC & Alternative Classifier\tabularnewline
\hline 
ACCD & Associative Classification over Concept Drifting Data Streams\tabularnewline
\hline 
ACDDM & Accurate Concept Drift Detection Method \tabularnewline
\hline 
ADDM & ADaptive sliding window based Detection Method\tabularnewline
\hline 
ADDS & Anti-concept Drift Detection Algorithm \tabularnewline
\hline 
ADWIN & ADaptive WINdowing \tabularnewline
\hline 
AGE & Accuracy and Growth rate updated Ensemble\tabularnewline
\hline 
ALD & Approximate Linear Dependence\tabularnewline
\hline 
AUE & Accuracy Updated Ensemble\tabularnewline
\hline 
AWE & Accuracy Weighted Ensemble\tabularnewline
\hline 
CDTs & Change Detection Tests\tabularnewline
\hline 
CSDD & Cosine Similarity Drift Detector\tabularnewline
\hline 
CUSUM & CUmulative SUM\tabularnewline
\hline 
DDD & Diversity for Dealing with Drifts \tabularnewline
\hline 
DDM & Drift Detection Method\tabularnewline
\hline 
DDM-OCI & Drift Detection Method for Online Class Imbalance\tabularnewline
\hline 
DELM & Dynamic extreme learning machine\tabularnewline
\hline 
DOED & Diversified Online Ensembles Detection\tabularnewline
\hline 
DWM & Dynamic Weighted Majority\tabularnewline
\hline 
ECDD & EWMA for Concept Drift Detection \tabularnewline
\hline 
ECHO & Efficient Concept Drift and Concept Evolution Handling over Stream
Data\tabularnewline
\hline 
ECPF & Enhanced Concept Profiling Framework\tabularnewline
\hline 
EDDM & Early Drift Detection Method \tabularnewline
\hline 
EDIST & Error DISTance for drift detection and monitoring \tabularnewline
\hline 
ELM & Extreme learning machine\tabularnewline
\hline 
ESOS-ELM & Ensemble of online sequential extreme learning machine\tabularnewline
\hline 
EWAUC & Equal Weighted AUC\tabularnewline
\hline 
EWMA & Exponentially Weighted Moving Average\tabularnewline
\hline 
FDR & False Discovery\tabularnewline
\hline 
FHDDM & Fast Hoeffding Drift Detection Method\tabularnewline
\hline 
FHDDMS & Stacking Fast Hoeffding Drift Detection Method\tabularnewline
\hline 
FHDDMSadd & Additive Stacking Fast Hoeffding Drift Detection Method\tabularnewline
\hline 
fnr &  False negative rate\tabularnewline
\hline 
FPDD & Fisher Proportions Drift Detector\tabularnewline
\hline 
fpr &  False positive rate\tabularnewline
\hline 
FSDD & Fisher Square Drift Detector\tabularnewline
\hline 
FsNB & Fast switch Na\i ve Bayes model\tabularnewline
\hline 
FTDD & Fisher Test Drift Detector\tabularnewline
\hline 
FTRL & Follow the Regularized Leader\tabularnewline
\hline 
FTRL-ADP & Follow-the-Regularized-Leader with Adaptive Decaying Proximal\tabularnewline
\hline 
HDDM & Hoeffding Drift Detection Method\tabularnewline
\hline 
HDWM & Heterogeneous Dynamic Weighted Majority\tabularnewline
\hline 
HLFR & Hierarchical Linear Four Rates\tabularnewline
\hline 
HT & Hoeffding Tree\tabularnewline
\hline 
KME & Knowledge-maximized ensemble\tabularnewline
\hline 
KS &  Kolmogorov-Smirnov test\tabularnewline
\hline 
LFR & Linear Four Rates\tabularnewline
\hline 
MD3 & Margin Density Drift Detection\tabularnewline
\hline 
MDDM & McDiarmid Drift Detection Method \tabularnewline
\hline 
meta-RRKOS-ELM-DDM  & Meta-cognitive Recurrent Recursive Kernel Online Sequential Extreme
Learning\tabularnewline
\hline 
MOS-ELM & Meta-cognitive online sequential extreme learning machine\tabularnewline
\hline 
NB & Naive Bayes\tabularnewline
\hline 
NDE & Number and Distance of Errors \tabularnewline
\hline 
NSE & Non stationary environments\tabularnewline
\hline 
OAUE & Online Accuracy Updated Ensemble\tabularnewline
\hline 
ODKK & Online drift detector for a K-class problem\tabularnewline
\hline
OMR-DDM & Online Map-Reduce Drift Detection Method\tabularnewline
\hline 
OS-ELM & Online sequential extreme learning machine\tabularnewline
\hline 
OWE & On-line Weighted Ensemble\tabularnewline
\hline 
PAUC & Prequential Multi-Class AUC\tabularnewline
\hline 
PHT & Page-Hinkley test\tabularnewline
\hline 
PINE & Predictive and Parameter INsensitive Ensemble\tabularnewline
\hline 
PPV & Positive Predictive Value\tabularnewline
\hline 
PSO & Particle Swarm Optimization\tabularnewline
\hline 
RDDM & Reactive Drift Detection Method\tabularnewline
\hline 
RDWM & Recurring Dynamic Weighted Majority\tabularnewline
\hline 
SEA & Streaming Ensemble Algorithm\tabularnewline
\hline 
SPC &  Statistical Process Control\tabularnewline
\hline 
STEPD & Statistical test of equal proportions\tabularnewline
\hline 
SVM & Support Vector Machines\tabularnewline
\hline 
TDAP &  Time Decaying Adaptive Prediction\tabularnewline
\hline 
tnr &  True negative rate\tabularnewline
\hline 
tpr &  True positive rate\tabularnewline
\hline 
UCEM & Uncertainty Error Correlation Matrix \tabularnewline
\hline 
WAC & Weighted AUC\tabularnewline
\hline 
WELM & Weighted extreme learning machine\tabularnewline
\hline 
WMA &  Weighted Majority Algorithm\tabularnewline
\hline 
WSTD & Wilcoxon Rank Sum Test Drift Detector\tabularnewline
\hline 
\end{longtable}
\endgroup

\section{Introduction}

In most real-world application scenarios, the machine learning model's performance deteriorates in production and consistently degrades as the systems evolve. This problem is commonly referred to as \textit{model degradation}. The accuracy of machine learning systems is prone to drop for various reasons. One reason could be that the data points on which the model was trained are not sufficient to capture the complexity of the problem space. Therefore, the model will perform unexpectedly for samples in the input space that was not covered in the instance space of training examples \cite{Marcus2018Deep, Weiss2004Rarity}. Another reason is that the system environment is dynamic and progressively subject to changes, making it difficult for a single model to provide accurate predictions. 

In the literature, researchers have distinguished between two main types of system changes concerning their nature. The first type is caused by changes of unknown context that cannot be measured or represented in the available attributes of the dataset, which is known as \textit{hidden context} \cite{Widmer1996HiddeCon}. Predictive systems typically struggle to cope with changes in hidden contexts, where an adaptive strategy is necessary to be executed. To illustrate the concept of hidden context, suppose a learning system should predict the Earth's temperature using only spatial and temporal historical data. Over time, the predictions will become inaccurate due to overlooking climate change that serves as a change in the hidden context, which is inaccessible information from the learner's view. Characterizing hidden context is generally domain-dependent, and in most cases, it cannot be expressible in such forms that benefit the learner if it would be incorporated. Therefore, researchers have extensively examined the second type of change, which is diagnosed in the underlying generating function of the data. This phenomenon is known as \textit{concept drift} \cite{Hoens2011Imbalance}. 

Concept drift might be attributed to changes such as degradation in the quality of materials of the system's equipment, seasonality, changing personal preferences and behaviors, or adversarial activities \cite{Barros2019Ens}. Since these sources of change are inherent elements of diverse real-world domains, concept drift has been introduced and addressed in a vast range of disciplines and domains. Recent applications include, but are not limited to,  IoT systems \cite{Mohsen2021Sensor, Xu2020IOT}, smart grids \cite{Fenza2019SG, Mohammadpourfard2020SG}, 5G networks \cite{Perepu20205G} and stock market \cite{Hu2015Stock, Cetrulo2019Market}. A recent study also has investigated the impact of concept drift on early alert systems during the SARS-CoV-2 pandemic \cite{Xu2021Pandemic}. These explored systems share the non-stationarity property since they are characterized by continuous changes as they develop. A broad overview of concept drift applications can be found in  \cite{Zliobaite2016Apps}. 

In the last decades, \textit{learning in non-stationary environments} \cite{Ditzler2015LNN} has been intensively studied. Researchers pointed out the necessity to integrate a model degradation detector in the overall learning framework. After deployment, the detector evaluates and tracks the system's performance to control such degradation in prediction accuracy. The error rate's degradation level is then used to signal concept drift alerts in the system.

Numerous terms and multiple mathematical definitions can be found in the literature to describe the same concept drift type. This lack of unified terminology in the field makes it challenging for researchers to find the correct definition of a given concept drift type. This paper investigates the terms and definitions used to describe the various types of concept drift. We also present a rigorous summary of concept drift and use a novel hierarchical classification. We categorize the existing approaches that explicitly rely on monitoring the error rate of the base learner to detect concept drift in the system. Incorporating such detection components in the system boosts the robustness of machine learning systems against changes and helps prevent the performance degradation of predictive models in our constantly changing world.

This paper is designed to address the following research questions:\\
\textbf{RQ1:} What are the different terms that are used in the literature to describe the same type of concept drift, and what are the same mathematical definitions used to describe the same concept drift type?\\
\textbf{RQ2:} What are the performance-based drift detection methods that were proposed in the last decade, and how can they be represented through a hierarchical classification?\\
\textbf{RQ3:} How is the model's predictive performance validated and used to track and detect concept drift, and what are the most common techniques utilized in the reviewed methods?\\

For \textbf{RQ1}, we delve into the literature to obtain the various terms that are used to refer to each concept drift type. For \textbf{RQ2} and \textbf{RQ3}, we survey the proposed performance-based concept drift detection methods in the last decade. From 2011 up to 2020, as illustrated later, researchers have principally extended or got inspired from the benchmark methods proposed in the preceding decade. The extended methods have primarily focused on improving the benchmark methods or enhancing their capability in dealing with more complex problems that involve fast-moving volumes of big data streams. We have chosen to review the works in the last decade since they can be viewed as the representatives of the most recent methods developed and compile the latest research gaps in the field.

The rest of this paper is organized as follows. Section \ref{Background} provides general background on drift detection and the related work. Section \ref{search} presents the search strategy that was implemented to address the research questions. Section \ref{Terminology} introduces the terms and definitions that are used in concept drift handling frameworks. Section \ref{Detectors} categorizes the performance-based concept drift detectors and reviews the existing approaches in the literature. An in-depth analysis and discussion on the surveyed methods are presented in Section \ref{analysis}. In Section \ref{conclusions}, we conclude the paper by presenting the main finding of this study and identifying future research directions.

%Bestoun --- 29 Aug 2021 ----1

\section{Background and Related Work}\label{Background}

\textit{Drift detection} or \textit{change detection} refers to the methodology that helps determine and identify a time instant or time interval when a change arises in the properties of the target object \cite{Basseville1993abrupt}. This definition has been extended to impose time constraints on the detection delay to enable the learner to adapt to the change efficiently to ensure high-performance \cite{Pears2014DDS}. 

Concept drift detection is a component of the concept drift handling framework that activates the \textit{concept drift adaptation} component, which reacts to the change in the data stream \cite{Lu2016LearningUnder}. Subsequently, the system will update the prior knowledge and adjust the learning models to react to the changes properly. This update usually raises the conflicting problem known as \textit{stability-plasticity dilemma} \cite{Grossberg1988StabPlas}. Here, stability means maintaining the relevant and possibly reoccurring knowledge. At the same time, plasticity describes replacing outdated knowledge in response to the new experience. Ideally, the concept drift solution should achieve a balance between stability and plasticity \cite{Elwell2011NSE}. Such concept drift adaptation strategy is referred to as \textit{informed} adaptation, or \textit{active} approach, which is triggered  upon drift occurrence detection to update the model. The other strategy is \textit{blind} adaptation, also denoted as \textit{passive} approach, where the model is constantly updated upon receiving new data instances without detecting drifts \cite{lobo2021Lunar, Song2021DAR}. 

\begin{figure}
\centering\includegraphics[width=0.7\linewidth]{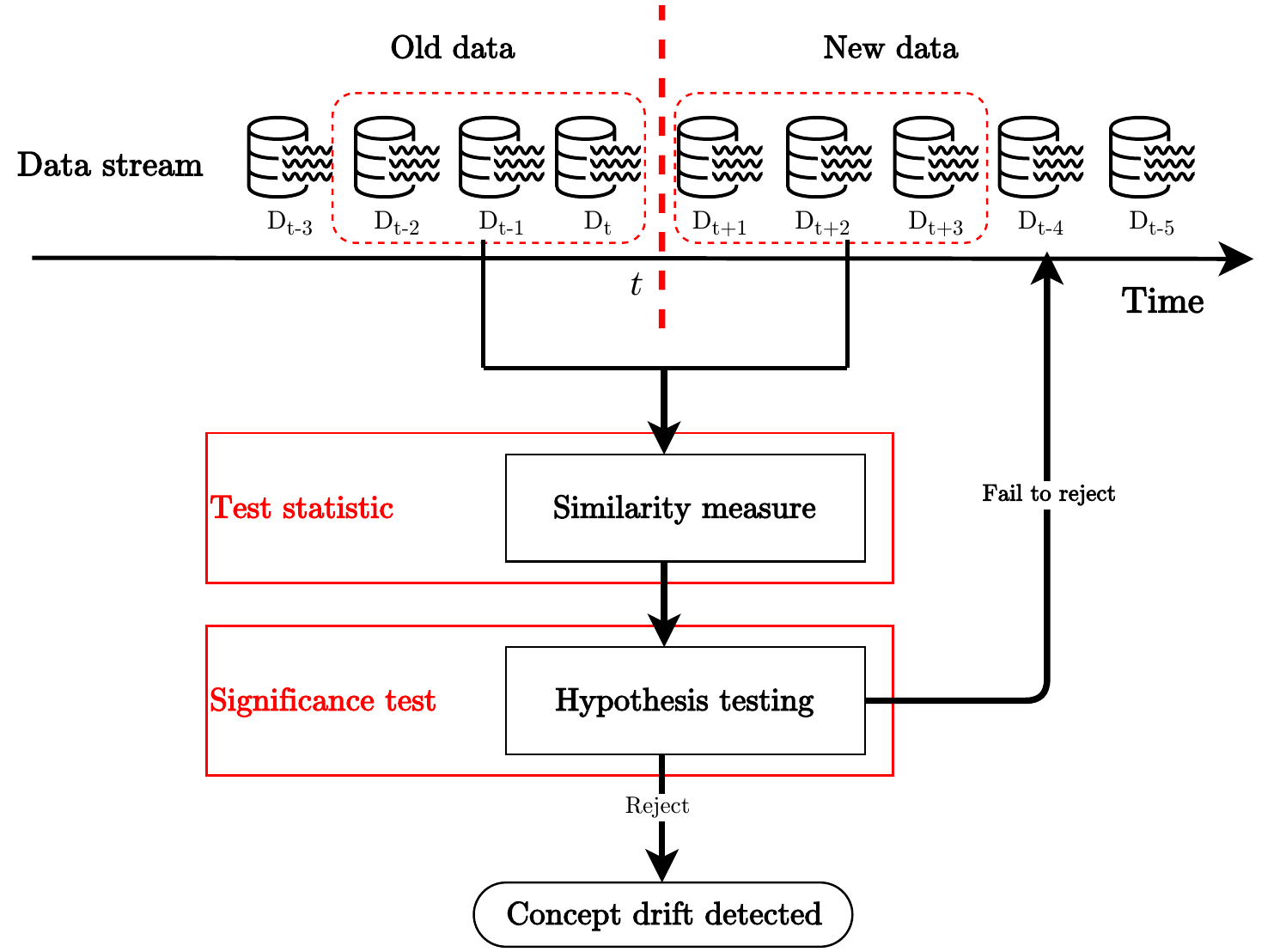}
\caption{Concept drift detection framework.}
\label{detection_test}
\end{figure}

Concept drift detection methods generally use a test statistic to keep tabs on the data stream and quantify the similarity between the old samples and the new ones to discern the change in the concept. This similarity value is then compared with a pre-defined threshold to find out the drift magnitude \cite{Dries2009AD}. Inspired by \cite{Lu2016LearningUnder}, Figure \ref{detection_test} summarizes a generic scheme for concept drift detection methods. In the figure, the null hypothesis is that the test statistic will not yield a significant difference between the old and new data, i.e., no concept drift detected. If failing to reject the null hypothesis, the system will persist with the current learner and slide on the data stream.

%Bestoun --- 30 Aug 2021 ----2

\begin{figure}
\centering\includegraphics[width=0.9\linewidth]{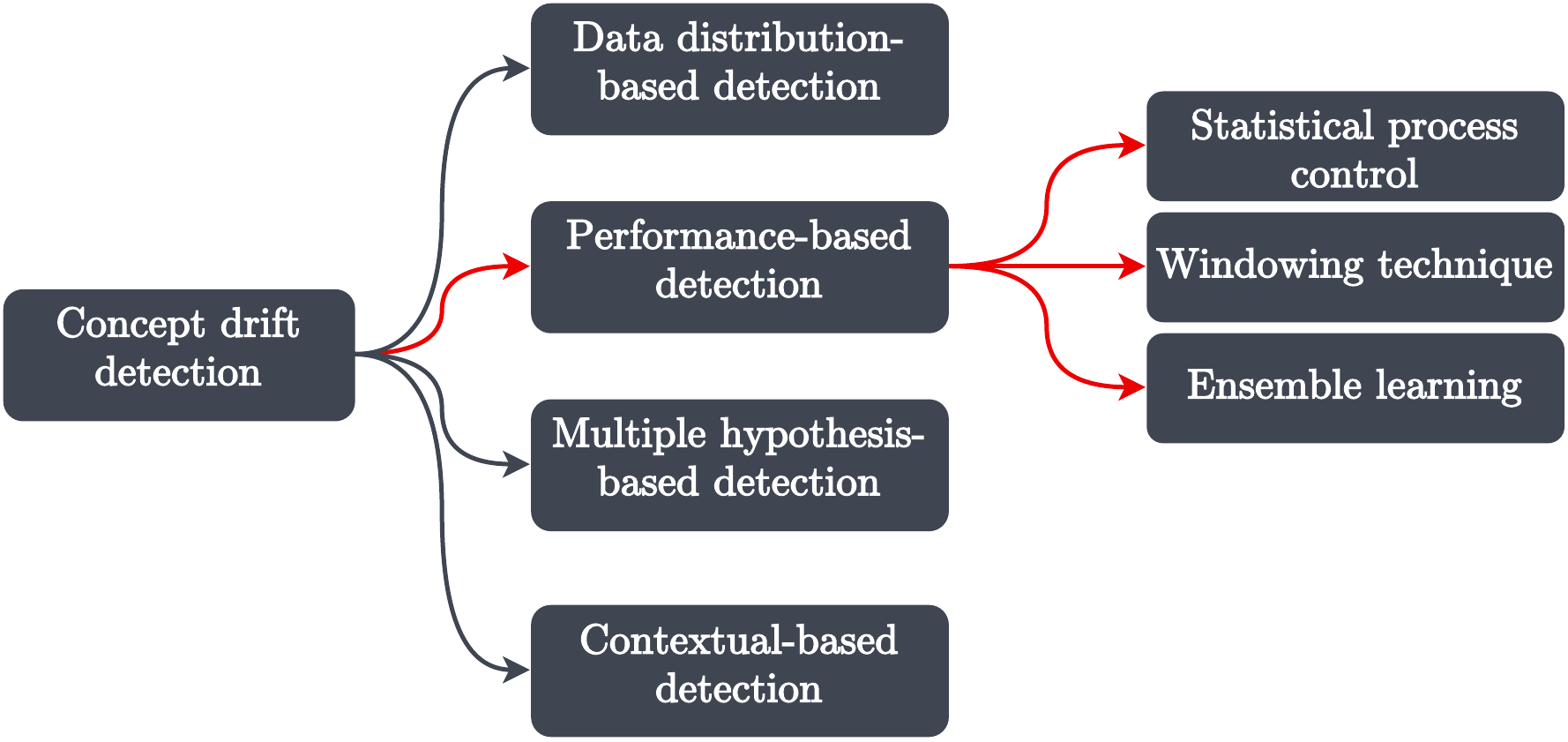}
\caption{Concept drift detection methods.}
\label{classification}
\end{figure}

Existing studies on detecting concept drift can be classified into different categories concerning the test statistics they apply to check and locate the change (see Figure \ref{classification}). Data distribution-based and performance-based, or error rate-based, approaches are the most dominant techniques used to detect concept drift since they can be applied to most learning tasks with lower complexity. There are also hybrid and contextual-based approaches.

%In Figure \ref{classification}, the red line highlights the research area of interest, performance-based approaches. This area has received substantial attention in the last decade and a considerable number of publications have been presented that we will survey and classify in this paper.

Data distribution-based detectors use distance measures to estimate the similarity between the data distributions in two different time-windows \cite{Kifer2004DCDStream}. Concept drift is then detected if the two distributions are significantly distant. Goldenberg and Webb \cite{Goldenberg2018Distance} summarize the distance measures that are used to compare the data distributions and estimate drifts. The main advantage of this approach is that it can be applied to both labeled and unlabeled datasets since this method only considers the distribution of data points. However, as we will discuss later, changes in the data distributions do not always affect the predictor performance, potentially leading to false alarms in the system \cite{Gama2014Adaptation}.

\begin{figure}
\centering\includegraphics[width=0.7\linewidth]{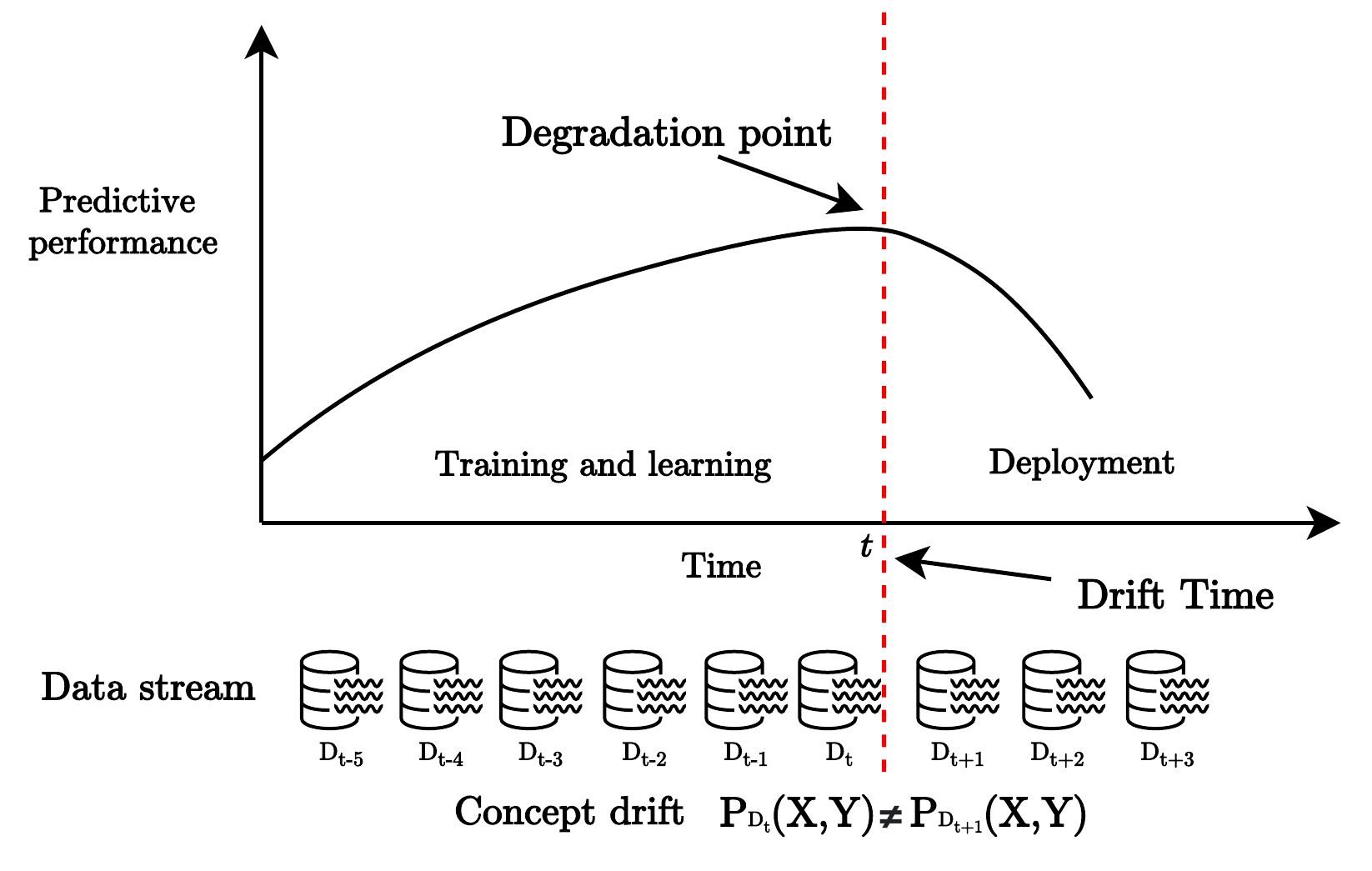}
\caption{Performance-based approach mechanism.}
\label{degradation_point}
\end{figure}

Performance-based approaches (as illustrated by the red arrows in Figure \ref{classification}) comprise the largest group of concept drift detectors. Therefore, they are the main focus of this paper for surveying and classification. These approaches typically trace deviations in the online learner's output error, known as the predictive sequential (prequential) error \cite{Gama2013Evaluating}, to detect changes \cite{Sebastio2009Study}. The basic idea of performance-based approaches aligns with \textit{Probability Approximately Correct} (PAC) learning model \cite{Mitchell1997ML}, which articulates that the prediction error depends on the size of the examples and the complexity of the hypothesis space. It concludes that if the examples are drawn from a stationary distribution, the error rate decreases as the learner sees more examples \cite{Gama2004DDM}. Thus, such a consequential decrease in the performance implies that the learned relationship between the examples of input data and the concept under study is obsolete, resulting in concept drift. Figure \ref{degradation_point} illustrates the main idea of performance-based approach mechanisms. Concept drift occurs when the joint distribution of the dataset $P_D(X,Y)$ changes at time instance $t$, which is the drift time. The main advantage of performance-based approaches is that they only handle the change when the performance is affected. Thus, these methods are more efficient in dealing with potential false alarms than distribution-based algorithms. However, the main challenge is that these methods require a quick arrival of feedback on the predictions, which is not always available \cite{Gama2014Adaptation}. Because of the limitation mentioned above, a new family of methods that deal with concept drift detection in unsupervised settings has been proposed.

Multiple hypothesis-based drift detectors are hybrid approaches that apply several detection methods and aggregate their results in parallel or hierarchically \cite{Lu2016LearningUnder}. Parallel drift detectors integrate the decisions of multiple drift detectors to make the final judgment. Hierarchical drift detectors incorporate two layers for drift detection. The first layer is the warning layer to alert the system about a potential occurrence of concept drift. The second layer is the validation layer that confirms or rejects the warning signaled from the first layer.

Contextual-based detectors use context information available from the system and data to detect the drift. Lu \textit{et al.} \cite{Lu2014Competence} have introduced concept drift detectors in a case-based reasoning system by tracking changes in competence measurement. Demšar and Bosnić \cite{Jaka2019exp} have used model explanation methodologies to interpret, visualize and detect concept drift. Lobo \textit{et al.} \cite{Lobo2018Spiking} have presented the eSNN-DD method that detects concept drift by exploiting the evolution of spiking neural networks. Huang \textit{et al.} \cite{Huang2015Volatility} have designed a concept drift detector using historical drift trends to calculate the probability of expecting a drift using online and predictive approaches. Graph metrics have also been utilized to detect concept drift in data streams that could be represented as graph streams as in \cite{Seeliger2017Graph, Paudel2020Graph, Zambon2018Graph}.

Several survey papers to formalize and classify concept drift were presented in the literature. In 2014, the most referenced survey on concept drift was published by Gama \textit{et al.} \cite{Gama2014Adaptation}. It covered and categorized concept drift handling systems from different perspectives and provided an excellent introduction to adaptive learning and concept drift. Another review paper \cite{Lu2016LearningUnder} summarized the research advancements on concept drift and proposed a new component in the concept drift handling framework, called \textit{concept drift understanding}. Ditzler \textit{et al.} \cite{Ditzler2015LNN} surveyed the studies on concept drift approaches from two main aspects, active and passive. Other related review studies \cite{Hu2020NoFLT,Khamassi2018DiscussionAR, Wares2019DSM, Iwashita2019Overview} have also surveyed and categorized the existing concept drift handling approaches. They have also provided an insightful discussion on the methods. Besides these review papers in the literature, other papers explored handling concept drift in specific learning tasks. A recently published review paper by Gemaque \textit{et al.} \cite{Gemaque2020Unsupervised} provides a full-scale overview of the methods that handle concept drift in unsupervised learning. Other papers review and scrutinize the progress in class-imbalanced data streams \cite{Wang2018Imbalance, Hoens2011Imbalance}. Krawczyk \textit{et al.} \cite{Krawczyk2017Ensemble} focuses on analyzing the research in ensemble learning for data streams in dynamic environments. However, with the availability of detailed review papers in concept drift classification and formalization, only a few studies have explored the different terms used by authors to describe the same type of concept drift, whereas new terms have appeared since the date of the publication of these studies. Additionally, far too little attention has been paid to performance-based concept drift detection. To this end, one of the main contributions of this paper is to narrow down the focus on performance-based concept drift detection approaches and provide a comprehensive summary of the recent progress in this research area, where not many review studies are available.

\section{Search Methodology}\label{search}

\begin{figure}
\centering
\hspace*{-1.5cm}
\includegraphics[width=1.2\linewidth]
{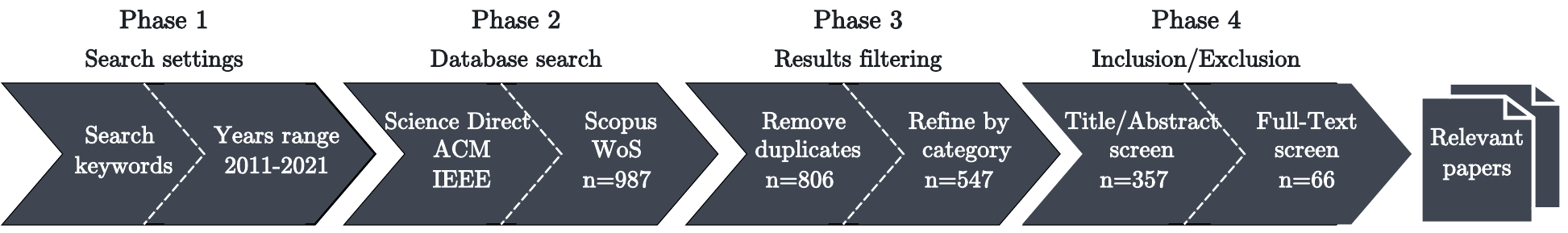}
\caption{Search strategy implemented to retrieve the relevant papers in the literature}
\label{search_method}
\end{figure}

\begin{figure}
\centering
\hspace*{-1.5cm}
\includegraphics[width=1.2\linewidth]
{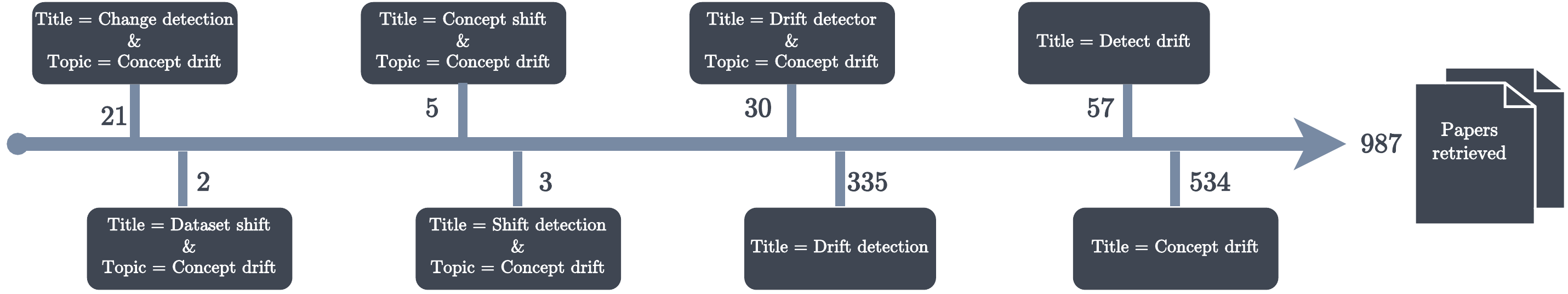}
\caption{Search terms for retrieving the papers}
\label{search_terms}
\end{figure}

The main objective of this paper is two-fold. First, the paper formalizes the problem of concept drift and surveys the terminologies used in the literature to describe its types, which is cataloged in \textbf{RQ1}. Second, it reviews the recent studies and trends in performance-based concept drift detection, which is addressed in \textbf{RQ2} and \textbf{RQ3}. Since the field has started to materialize in the early 2000s, we decided to retrieve the terms that have appeared in the studies of the last two decades while we retrieved the performance-based detection methods of the last decade. The main reason why we have followed different methodologies in addressing \textbf{RQ1} and, \textbf{RQ2} and \textbf{RQ3}, is that most of the terms have appeared in the early aughts of the present century and have been used afterward by the authors. In contrast, we limited the retrieval of the detection methods to the last decade to determine the current trends in the research area. 
\linebreak To address \textbf{RQ1}, we have explored the terms used in former surveys and through snowballing of highly-cited references. For \textbf{RQ2} and \textbf{RQ3},
and while this paper does not directly follow a systematic literature review protocol, we have followed a systematic literature search methodology to retrieve and select relevant papers that answer \textbf{RQ2} and \textbf{RQ3}, as shown in Figure \ref{search_method}. In the first phase, we define the search settings. We set the date range to the last decade (2011-2021) and defined the keywords to use for our search queries since concept drift appeared under different terminologies in the literature. We have summarized the search terms that we used in Figure \ref{search_terms}. In phase 2, we set the paper index database resources we searched to retrieve the papers. We acquired the papers from the top database indices, including IEEE Xplore, Science Direct, ACM, Scopus, and Web of Science. This resulted in 987 publications. In Phase 3, we filtered the results by removing duplicates, resulting in 806 papers. In phase 4, we refine the search results and constrain them to the studies published in the computer science discipline. In addition to the search phases, we performed screening to segregate the relevant papers. We carry out the screening in two levels. The first level investigates the title and abstract to exclude the papers that do not detect concept drift. Then, we scrutinize the full text in the second level to include the relevant studies that use the model's performance to detect concept drift.

%Bestoun --- 3 Sep 2021 ----5

To decide whether to include or exclude the paper, we have considered a set of inclusion/exclusion criteria to determine relevant publications. 
\begin{enumerate}
  \item We removed papers that are not published in English.
  \item The paper must be peer-reviewed according to the formal peer-review process in the scientific community. That filters out preprints, book chapters, Master or Ph.D. dissertations.
  \item Survey papers were excluded, since they do not introduce a new concept drift detection approach.
\end{enumerate}

To determine the approaches relevant to the scope of this paper, we selected the candidate papers according to the following inclusion criteria:
\begin{enumerate}
  \item The approach must propose a novel drift detection method or integrate existing drift detectors in new predictive systems.
  \item The approach must be general and not only targeted to solve a specific problem or installed in a particular domain or application.
  \item The approach must explicitly detect concept drift.
  \item The approach must use the learner's performance to detect the drift without the underlying data distribution.
\end{enumerate}

Following the above criteria, 66 papers remained to be reviewed for addressing \textbf{RQ2} and \textbf{RQ3}. The following sections present the result of our analysis and address the research questions.

\section{Terminology and Definitions}\label{Terminology}
As mentioned previously, researchers have defined and mathematically represented concept drift and its derivatives in different ways. In 2012, Moreno-Torres \textit{et al.} \cite{Moreno2012DShift} first addressed this lack of standard terminology and suggested that concept drift is a type of the generic phenomenon \textit{dataset shift} that covers \textit{covariate shift}, \textit{prior probability shift} and \textit{concept drift}. Each concept drift type is framed by a certain change in the data distribution. But since the date of this publication, new terms have appeared in the literature, and novel concept drift types have emerged. 

To address \textbf{RQ1}, this section will provide a taxonomy to group the various terms used in the literature, starting from mathematical definitions of each variant. The taxonomy will help the researchers and practitioners to gain a unified and consolidated view on the notations by providing precise and concise terminology in the field. In the following subsections, we will formally define concept drift and the different types and survey the terms used to describe each type.

%Bestoun --- 30 Aug 2021 ----3

\subsection{Notation and Formalism for Concept Drift}

In supervised machine learning tasks, each data instance is defined by a pair of feature vectors or covariates $X$, and a target variable or response $y$. \cite{Gama2014Adaptation} have introduced a probabilistic definition to describe a time-varying \textit{concept} as the joint distribution of $X$ and $y$ at time $t$, $P_t(X, y)$. Tracking a change in data samples requires a time-ordered sequence of instances. Concept drift is usually aligned in the \textit{stream learning} context since a data stream is defined as a continuous, potentially unbounded, sequence of data elements with associated time stamps arriving in sequential order \cite{Gama2010KDDS}. This is in contrast to dataset shift in a \textit{batch learning} scenario, where the data is entirely stored in memory and processed all at once \cite{Wares2019DSM}. Changes are characterized between the training and testing probability distributions \cite{Candela2009DatasetS}. Thus, concept drift is viewed as the stream learning correspondent of the dataset shift in the batch setting \cite{Webb2016Chara}. 

Concept drift is formally defined as a change in the joint distribution between two time instances $t$ and $t+w$ \cite{Klinkenberg2004Example}, where $t$ could be a particular time point or time interval, and $w$ denotes the time window when the distribution change is being checked at. Consequently, concept drift occurs, if:

\begin{equation}
    P_{t} (X,y) \neq  P_{t+w} (X,y)
    \label{eq1}
\end{equation}

In a recent study \cite{Song2021SEGA}, authors suggested adding an extra constraint to the definition presented in Eq.\ref{eq1} to guarantee that the new concept will retain for some time period (at least for two time points):
\begin{equation}
    \forall i, \quad w = \tau_{d^{(i+1)}}-\tau_{d^{(i)}}>1
    \label{cons1}
\end{equation}
where $\tau \in \mathbb{Z}^{+}$ is the time point, and $d^{(i)}$ denotes the time point order of the $i$th concept drift appeared in the system. This additional constraint will distinguish concept drift from outliers that last momentarily and ensures that the concept drift is a new pattern rather than an ephemeral disturbance in the data (i.e., noise).

Starting from the product rule, and according to the Bayesian Decision Theory \cite{Duda2001Pattern} the joint distribution in Eq.\ref{eq1} can be decomposed and rewritten as:
\begin{equation}
    P_t(X,y) = P_t(y|X) \times P_t(X) = P_t(X|y) \times P_t(y)
    \label{eq2}
\end{equation}

In the settings of classification problems,
   \begin{itemize}
     \item $P_t(y|X)$ denotes the posterior probability distribution of the target labels, 
     \item $P_t(X)$ is the input data probability distribution,
     \item $P_t(y)$ denotes  the prior probability distribution of the target labels,
     \item $P_t(X|y)$ denotes the class-conditional probability density distribution.

   \end{itemize}

\subsection{Concept Drift Types}

Researchers categorized concept drift into different types in terms of the form that it takes place in the system. The probabilistic source of change and the arrival pattern (i.e., drift transition) are the most commonly used principles to distinguish concept drift. There are also other criteria to categorize concept drift, such as speed, severity, and recurrence. \cite{Webb2016Chara, Khamassi2018DiscussionAR} present an exhaustive categorization of concept drift types. The following subsections provide a comprehensive taxonomy of concept drift types, categorized by the probabilistic source of change and drift transition.

\subsubsection{Probabilistic Source of Change}

\begin{figure}[!ht]
\centering\includegraphics[width=1\linewidth]{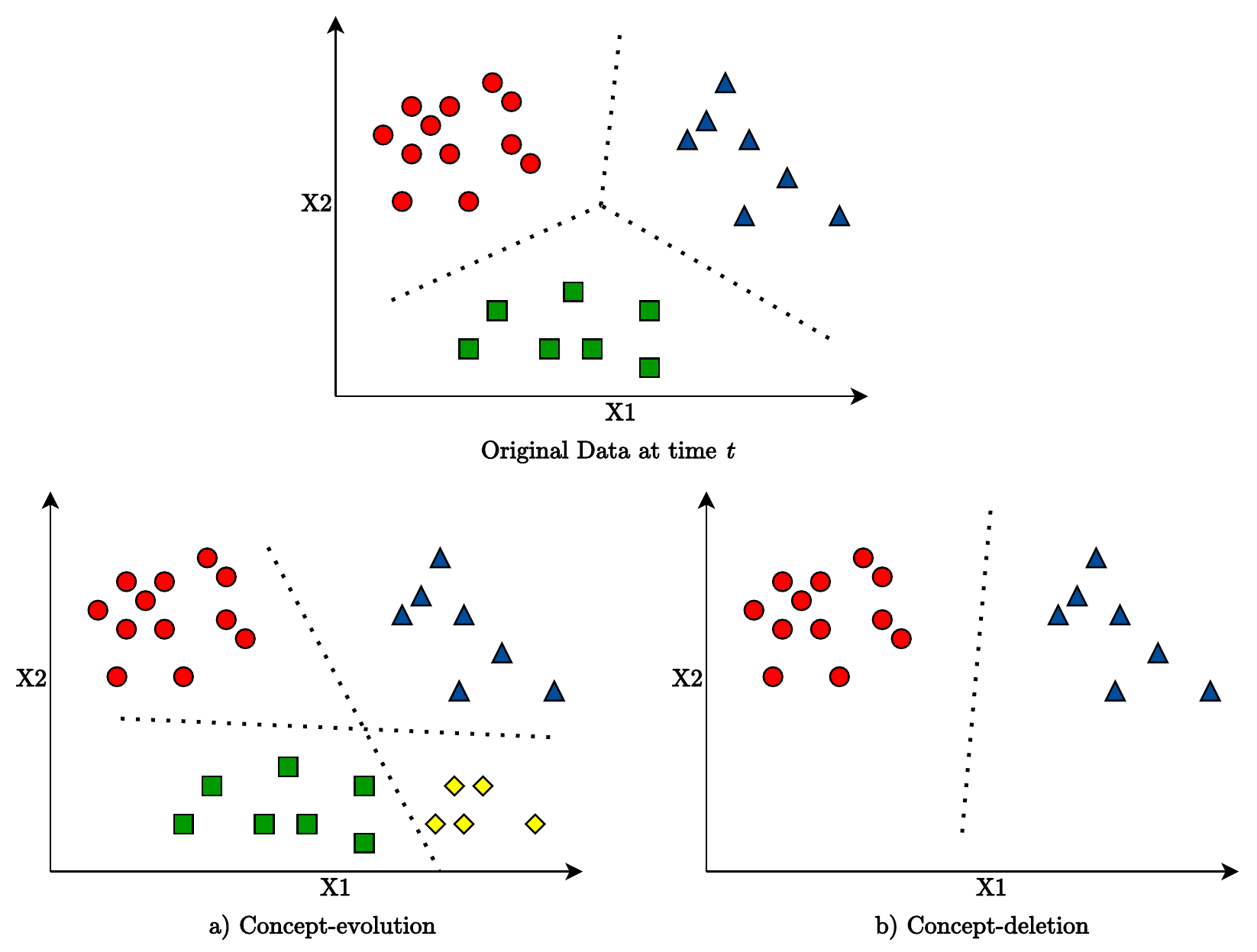}
\caption{Concept-evolution and concept-deletion}
\label{concept_evolution}
\end{figure}

This type of concept drift is the most closely studied in the literature. To make the inequality of Eq.\ref{eq1} hold, it identifies the changes in the probability distributions. As can be seen from Eq.\ref{eq2}, any concept drift type, assuming probability distribution change, is associated with at least another type since a change in any probability distribution in Eq.\ref{eq2} will induce at least one change in another distribution. To illustrate that argument, we consider the posterior probability distribution as an example. If $P_{t} (y|X) \neq  P_{t+w} (y|X)$ then, by applying the Bayesian rule:

\begin{equation}
    \frac{P_t(X|y) \times P_t(y)}{P_t(X)} \neq \frac{P_{t+w}(X|y) \times P_{t+w}(y)}{P_{t+w}(X)}
    \label{eq3}
\end{equation}
The inequality of Eq.\ref{eq3} holds if, at least, one of the probability distributions that compose it has changed. A similar argument can be used for the other probability distributions. The probabilistic sources of drift are then defined as follows:

\begin{enumerate}
  \item  $P_{t} (y|X) \neq  P_{t+w} (y|X)$:
  A change in the posterior probability distribution indicates a principal change in the underlying target concept. This drift type directly affects the prediction performance since it requires an adaptation of the decision boundary to react to it for preserving the model's accuracy. There are mainly two types, where this form of drift takes place. The first type is mainly referred to as \textit{real concept drift}, Figure \ref{drift_types}(a), where changes in $P(y|X)$ might or might not be associated with changes in $P(X)$ \cite{Gama2014Adaptation}. The second type manifests itself without a change in the data distribution $P(X)$. This type is called \textit{actual drift} \cite{Lu2016LearningUnder}, as illustrated in Figure \ref{drift_types}(b). This paper mainly covers real concept drift detectors since these methods detect drifts that affect the predictor's performance.

  \afterpage{%
\thispagestyle{empty}
\begin{figure}[!ht]
\centering
\vspace*{-3cm}
\includegraphics[width=0.9\linewidth]
{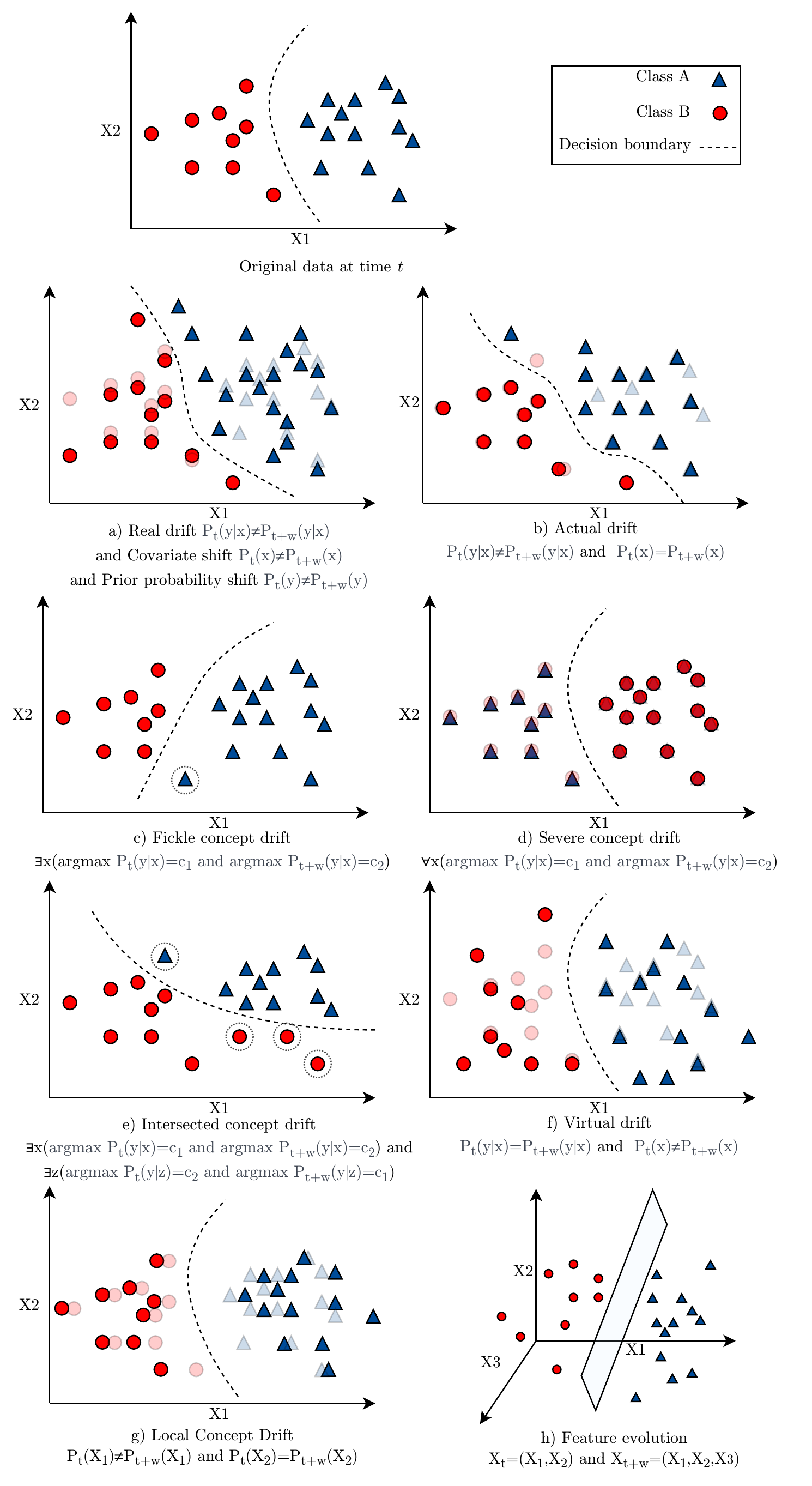}
\caption{Concept drift types by probabilistic source of change}
\label{drift_types}
\end{figure}
\clearpage
}

 %Bestoun --- 1 Sep 2021 ----4
 
  There are also subcategories derived from this probabilistic source of change. \textit{Fickle concept drift} occurs when some data samples belong to two different classes at two different times \cite{Forman2006Fickle}, which can be written mathematically as $\exists x (\textit{argmax } P_t(y|x)=c_1$ and $\textit{argmax } P_{t+w}(y|x)=c_2)$, Figure \ref{drift_types}(c). \textit{Severe concept drift} occurs if the target classes of all the data samples change after the drift occurrence \cite{Minku2010Diversity}. This type of drift is also called \textit{full-concept drift} \cite{Webb2016Chara}, Figure \ref{drift_types}(d). This type of concept drift can be represented mathematically as $\forall x (\textit{argmax } P_t(y|x)=c_1$ and $\textit{argmax } P_{t+w}(y|x)=c_2)$.
  \textit{Intersected concept drift} occurs when only a subspace of the data samples changes their target classes after the drift occurrence \cite{Minku2010Diversity}, which is also referred to as \textit{subconcept drift} \cite{Webb2016Chara}, Figure \ref{drift_types}(e). This type of concept drift can be represented mathematically as $\exists x (\textit{argmax } P_t(y|x)=c_1$ and $\textit{argmax } P_{t+w}(y|x)=c_2)$ and $\exists z (\textit{argmax } P_t(y|z)=c_2$ and $\textit{argmax } P_{t+w}(y|z)=c_1)$.
  \item $P_{t} (X) \neq  P_{t+w} (X)$:
  A change in the underlying data distribution is mainly referred to as \textit{covariate shift }\cite{Shimodaira2000Covariate}. Figure \ref{drift_types}(a) illustrates the covariate shift.
  If the input data distribution changes without affecting the target concept, and hence the decision boundary, it is called \textit{virtual drift} \cite{Ramirez2017Survey}, in mathematical terms, $P_{t}(y \mid x)=P_{t+w}(y \mid x)$ and $P_{t}(x) \neq P_{t+w}(x)$, Figure \ref{drift_types}(f). In practice, changes in the data and the posterior probability distributions often happen simultaneously \cite{Delany2005Spam}.
  
  \textit{Local concept drift} and \textit{Feature-evolution} are other subcategories that can be considered as of this probabilistic source of change. \textit{Local concept drift} refers to the situation where the distribution change targets only a sub-region of the feature space \cite{Tsymbal2008Local}. This can be expressed as $P_t(X_1) \neq P_{t+w}(X_1) $ and $P_t(X_2) = P_{t+w}(X_2) $, Figure \ref{drift_types}(g) illustrates the Local concept drift.  \textit{Feature-evolution} occurs when new attributes (e.g. $X_3$) dynamically arise in the input space \cite{Masud2010Evolution}, i.e when $X_t \neq X_{t+w}$, and as a result, $P_t(X) \neq P_{t+w}(X)$, where $X$ is the set of input variables, Figure \ref{drift_types}(h).

  \item $P_{t} (y) \neq  P_{t+w} (y)$
  A change of the distribution of classes over time is referred to as \textit{prior-probability shift} \cite{Candela2009DatasetS} as illustrated in Figure \ref{drift_types}(a). This drift type could affect the prediction performance if there is a significant change in the distribution of classes or the number of classes in the learning problem has changed.
  
Another subcategory of this source of change that is found in the literature is \textit{concept-evolution}, which refers to the emergence of novel classes in the problem \cite{Masud2010Evolution} as illustrated in Figure \ref{concept_evolution}(a). Similarly, \textit{concept deletion} refers to the disappearance of classes in the problem \cite{Elwell2011NSE}, Figure \ref{concept_evolution}(b).
\end{enumerate}

Table \ref{tab:distribution} summarizes the terms that can be found in the literature to describe concept drift types that are characterized by the probabilistic source of change. The terms in the table are also grouped by the corresponding mathematical definitions.

\subsubsection{Transition of Change}
\begin{figure}[!ht]
\centering
\includegraphics[width=1\linewidth]{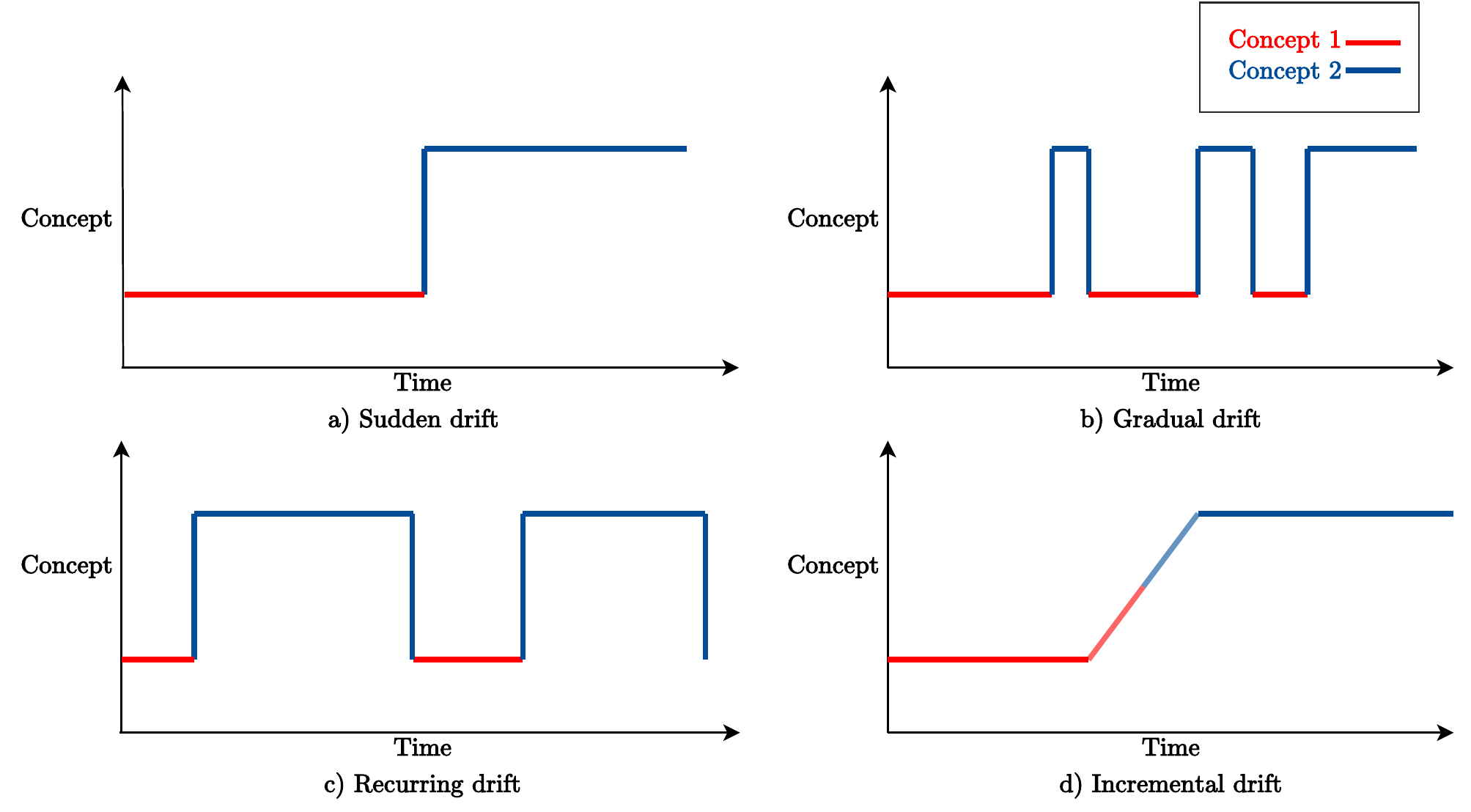}
\caption{Concept drift categorized by pattern of arrival}
\label{drift_transition}
\end{figure}

This categorization distinguishes concept drift characteristics based on the pattern of how the drift evolves in the system. It can be classified as follows \cite{Lu2016LearningUnder}: 
\begin{enumerate}
    \item \textbf{Sudden Drift:} Occurs when the target distribution changes from one concept to another abruptly at a point in time (e.g., Figure \ref{drift_transition}(a)).
    \item \textbf{Gradual Drift:} Occurs when the target distribution changes progressively from one concept to another (e.g., Figure \ref{drift_transition}(b)).
    \item \textbf{Recurring Drift:} Occurs when a precedently-seen concept reappears again after a time interval (e.g.,  Figure \ref{drift_transition}(c)). This type is similar to the gradual drift since the two concepts interchange in the system, but the main difference is the transition phase. In gradual drift, the old concept starts to phase out and is to be replaced with the new one increasingly. While in recurring drift, the old concepts reoccur after some time \cite{Ramirez2017Survey}.
    \item \textbf{Incremental Drift:} Occurs when a new concept replaces the old one slowly in a continuous manner (e.g., Figure \ref{drift_transition}(d)). Some Authors consider this type as a sub-type of gradual drift \cite{Krawczyk2017Ensemble}, as in the two drift types, the new concepts emerge in the system and completely replace the old one. While the difference is that in the incremental drift, there is no obvious boundary that separates the occurrence of the different concept \cite{Zhong2021Long}.

\end{enumerate}

Table \ref{tab:speed} summarizes the terms that can be found in the literature to describe the aforementioned concept drift types that are characterized by the transition of change.

Table \ref{tab:distribution} and Table \ref{tab:speed} answer \textbf{RQ1} by providing an overview of the different terms used by researchers to refer to concept drift types in the literature.

\aboverulesep=0ex  
\belowrulesep=0ex 
\renewcommand{\arraystretch}{1.3}
\strutlongstacks{T}
\renewcommand\thetable{\arabic{table}}

\begin{sidewaystable}

\vspace*{-0.5cm}

\begin{minipage}[b]{1\textwidth}
    \setcounter{table}{1} 
    \captionof{table}{Concept drift by probabilistic source of change}
    \label{tab:distribution}
\end{minipage}

\vspace*{-0.5cm}\hspace{-4.5cm}
\scalebox{0.88}{
\begin{minipage}{1\textwidth}
\begin{longtable}[c]{|c|c|c|c|c|c|c|}

\toprule 
\textbf{\Longstack{Mathematical \\ definition}} &        $P_{t} (X,y) \neq  P_{t+w} (X,y)$ & 
        $P_{t} (y|X) \neq  P_{t+w} (y|X)$ & 
        \shortstack{$P_{t} (y|X) =  P_{t+w} (y|X)$ \\and $P_{t} (X) \neq  P_{t+w} (X)$ } & \shortstack{$P_{t} (y|X) \neq  P_{t+w} (y|X)$ \\and $P_{t} (X) =  P_{t+w} (X)$ }&  
        $P_{t} (X) \neq  P_{t+w} (X)$ & 
        $P_{t} (y) \neq  P_{t+w} (y)$ \tabularnewline
\midrule
\hline
\endhead
\multirow{8}{*}{\textbf{\Longstack{Terminologies \\used}}} & \multirow{3}{*}{\text{\Longstack{Concept Drift\\\cite{Zhang2008Categorizing, Gama2014Adaptation, Webb2017UnderstandingCD}}}} & \Longstack{Real Concept Drift \\ \cite{Gama2014Adaptation, Kolter2007DWM, Syed1999ILSVM}}& \Longstack{Virtual Drift \\ \cite{Widmer1993ContextTrack, Tsymbal2004DefandRel}} &\Longstack{Actual Drift \\ \cite{FdezRiverola2007Lazy, Lu2016LearningUnder}} & \multirow{2}{*}{\Longstack{Covariate Shift \\ \cite{Shimodaira2000Covariate, Sugiyama2012IntroductiontoCS}}} & \multirow{2}{*}{\Longstack{Prior-Probability \\Shift\cite{Candela2009DatasetS, Moreno2012DShift} }}\tabularnewline
\cmidrule{3-5} \cmidrule{4-5} \cmidrule{5-5} 
 &  & Concept Drift\cite{Krempl2011Latency, Elwell2011NSE} & Temporary Drift\cite{Lazarescu2004Permenant} & \multirow{2}{*}{\Longstack{Conditional Change \\\cite{Gao2007AGF}}} &  & \tabularnewline
\cmidrule{3-4} \cmidrule{4-4} \cmidrule{6-7} \cmidrule{7-7} 
 &  & Concept Shift\cite{Salganicoff1997CShift, Gama2014Adaptation}     &Sampling Shift \cite{Salganicoff1997CShift, Tsymbal2004DefandRel}   &  & \multirow{2}{*}{\Longstack{Virtual Drift \\ \cite{Tsymbal2008Local}}} & Global Drift\cite{Hofer2013Global} \tabularnewline
\cmidrule{2-5} \cmidrule{3-5} \cmidrule{4-5} \cmidrule{5-5} \cmidrule{7-7} 
 & \multirow{3}{*}{\text{\Longstack{Dataset Shift \\ \cite{Candela2009DatasetS, Moreno2012DShift} }}} & Permanent Drift\cite{Lazarescu2004Permenant, Gama2010KDDS}  & Feature Change\cite{Gao2007AGF}  & Real Concept Drift\cite{Iwashita2019Overview}     &  & Label Shift\cite{Lipton2018LabelShift, Azizzadenesheli2019RegularizedLF}   \tabularnewline
\cmidrule{3-7} \cmidrule{4-7} \cmidrule{5-7} \cmidrule{6-7} \cmidrule{7-7} 
 &  & Conditional Shift \cite{Zhang2013Target, Subbaswamy2019Preventing}  & \multirow{2}{*}{\Longstack{Loose Concept \\ Drifting\cite{Zhang2008Categorizing}  }} & \multirow{3}{*}{\Longstack{Concept Shift\\ \cite{Moreno2012DShift, Heiser2020Shift} }} & \multirow{4}{*}{\Longstack{Data Distribution \\ Drift\cite{Sethi2016Grid}  }} & \multirow{2}{*}{\Longstack{Target Shift \\ \cite{Zhang2013Target, Nguyen2016Target}  }}\tabularnewline
 \cmidrule{3-3}
 &  & Rigorous Concept Drifting\cite{Zhang2008Categorizing}  &  &  &  & \tabularnewline
\cmidrule{2-3} \cmidrule{3-3} \cmidrule{3-4} \cmidrule{7-7} 
 & \multirow{2}{*}{\Longstack{Concept Shift\\ \cite{Moreno2012DShift, Vorburger2006Entropy} }} & Population Drift\cite{Kelly1999Population}  & \multirow{2}{*}{\Longstack{Pure Covariate \\ Drift\cite{Webb2016Chara} } } &  &  & \multirow{2}{*}{\Longstack{Class Prior Shift \\ \cite{Charoenphakdee2019ClassPrior, Hoens2011Imbalance}}}\tabularnewline
\cmidrule{3-3} \cmidrule{5-5} 
 &  & Class Distribution Drift\cite{Sethi2016Grid}  & & Pure Class Drift\cite{Webb2016Chara} &  & \tabularnewline
\bottomrule
\end{longtable}
\end{minipage}
}

\vspace{25ex}

\begin{minipage}[b]{1\textwidth}

    \setcounter{table}{2} 
    \captionof{table}{Concept drift by probabilistic source of change}
    \label{tab:speed}
\end{minipage}

\vspace*{-0.5cm}\hspace{-1.25cm}
\begin{minipage}{1\textwidth}

\centering
\begin{longtable}{|c|c|c|c|c|}
\toprule 
\text{\textbf{Primary Term}} & \text{\Longstack{Sudden Drift \\ \cite{Stanley2003Learning, Tsymbal2004DefandRel} }} & \text{\Longstack{Gradual Drift \\ \cite{Stanley2003Learning, Hickey2001Refined}}} & \text{\Longstack{Recurring Drift \\ \cite{Mauricio2013RCD, Jagadeesh2011Process} }} & \text{\Longstack{Incremental Drift\\\cite{Vzliobaite2010Learning, Jagadeesh2011Process}}}\tabularnewline
\hline
\endhead
\midrule 
\multirow{4}{*}{\text{\textbf{Alternative terms}}} & \text{Abrupt Drift\cite{Tsymbal2004DefandRel, Brzezinski2013AUE2}  } & \multirow{4}{*}{\text{\Longstack{Evolutionary Drift \\ \cite{Black1999Maintaining, Narasimhamurthy2007Framework}}}} & \multirow{2}{*}{\text{\Longstack{Recurring Contexts \\ \cite{Mauricio2013RCD, Katakis2010Recurring}}}} & \multirow{2}{*}{\text{\Longstack{Stepwise Drift \\ \cite{Vzliobaite2010Learning, Fabricio2013Detection}}}}\tabularnewline
\cmidrule{2-2} 
 & \text{Concept Shift\cite{Syed1999ILSVM, Vorburger2006Entropy}} &  &  & \tabularnewline
\cmidrule{2-2} \cmidrule{4-5} \cmidrule{5-5} 
 & \text{Revolutionary Drift\cite{Black1999Maintaining, Narasimhamurthy2007Framework}} &  & \multirow{2}{*}{\text{\Longstack{Replacing Drift \\ \cite{Yazdi2020UECM}}}} & \multirow{2}{*}{\text{\Longstack{Development Drift \\ \cite{Yazdi2020UECM}}}}\tabularnewline
\cmidrule{2-2} 
 & \text{Immediate Drift\cite{Yazdi2020UECM}} &  &  & \tabularnewline
\bottomrule

\end{longtable}
\end{minipage}
\end{sidewaystable}
\pagebreak

\section{Performance-Based Concept Drift Detectors}\label{Detectors}
This section surveys performance-based concept drift detection methods to answer \textbf{RQ2}. These methods can be categorized according to the strategy used to detect drops in performance: statistical process control, windowing techniques, and ensemble learning, as illustrated in Figure \ref{classification}. To guide the reader, we summarize the reviewed approaches in this paper as illustrated in Figure \ref{navigation}. The navigation diagram is based on a hierarchical scheme that connects the original method with its derivatives and extensions.

\afterpage{
\begin{figure}
\thispagestyle{empty}
\vspace{-1cm}\hspace{-2.6cm}
\includegraphics[width=1.35\linewidth]
{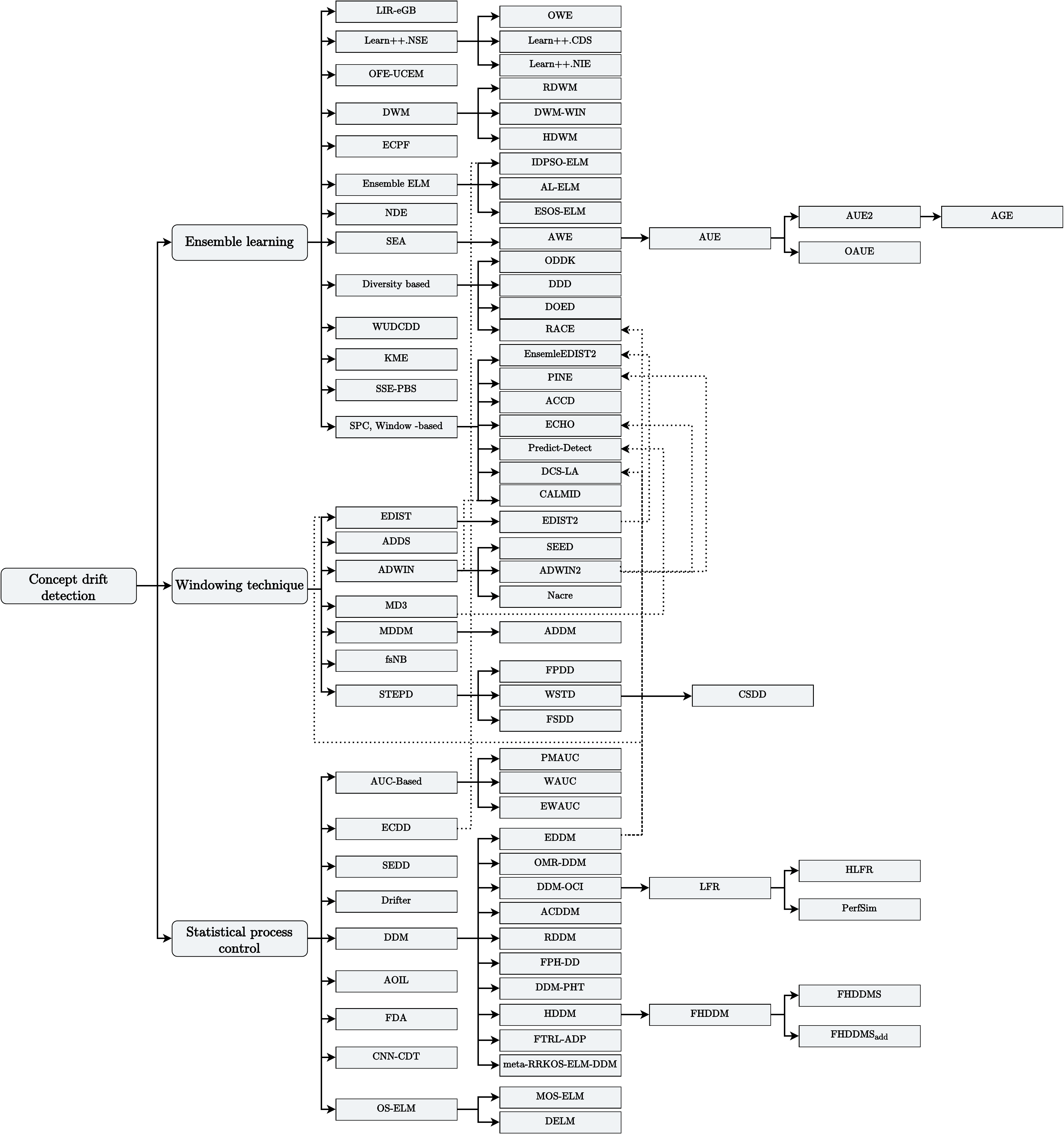}
\caption{Classification hierarchy of reviewed performance-based drift detector methods}
\label{navigation}
\end{figure}
\clearpage
}

\subsection{Statistical Process Control}
The Statistical Process Control (SPC) criterion is used to monitor the quality of the learning process by tracing the online error rate evolution of base learners. Concept drift is assumed to have occurred if the model's performance degradation exceeds the significance test level. Numerous performance-based methods can be found in the literature that rely on SPC to detect concept drift.

The Drift Detection Method (DDM) \cite{Gama2004DDM} is a well-known and widely-used algorithm and has been used as conceptual underpinning for a number of related performance-based drift detectors. DDM analyzes the error rate of the streaming data classifier to detect changes. The method considers the error as a Bernoulli random variable with Binomial distribution. It monitors $p_t$, the probability of misclassification at time $t$, and the standard deviation $s_{t}$ as:
\begin{equation}
   s_t = \sqrt{p_t(1-p_t)/i}
    \label{eq4}
\end{equation}

At time $t$, $p_{min}$ and $s_{min}$ are replaced with the corresponding values of $p_{t}$ and $s_{t}$, if $p_{t}+s_{t} < p_{min}+s_{min}$. The method defines a warning state which is triggered when  $p_{t}+s_{t} \ge p_{min}+2*s_{min}$, and a drift is detected when $p_{t}+s_{t} \ge p_{min}+3*s_{min}$.

Other methods have modified DDM to enhance its performance for solving diverse tasks. For example, Early Drift Detection Method (EDDM) \cite{Baena2006EDDM} extends DDM by tracking the distance between two consecutive misclassifications rather than the error rate. This approach was proven to be more efficient than DDM in detecting gradual drifts \cite{Nishida2007STEPD}. Reactive Drift Detection Method (RDDM) \cite{Barros2017RDDM} mitigates the performance loss problem of DDM, which is due to decreased sensitivity when the concept has a large number of members. RDDM augments DDM by periodically removing old data instances of long concepts. The authors argued that RDDM provides higher or equal global accuracy than DDM and detects drifts earlier in most situations. 

Hoeffding Drift Detection Method (HDDM) \cite{Frias2014HDDM} modifies DDM by using the Hoeffding’s inequality \cite{Hoeffding1963Probability} to detect substantial changes in the moving average of the performance estimate. The authors proposed two variants of the method, HDDM\textsubscript{A} that is suitable to detect sudden drifts, and HDDM\textsubscript{W} for gradual drifts. Fast Hoeffding Drift Detection Method (FHDDM) \cite{Pesaranghader2016FHDDM} addressed the shortcomings of HDDM caused by high numbers of false positives and false negatives. FHDDM employs a sliding window to compare the maximum overall probability of a correct prediction and the most recent one. Stacking Fast Hoeffding Drift Detection Method (FHDDMS) and Additive FHDDMS (FHDDMS\textsubscript{add}) \cite{Pesaranghader2018FHDDMS} extends FHDDM by maintaining windows of different sizes (short and long sliding windows) to detect various types of drift. FHDDMS\textsubscript{add} uses a binary indicator of classification errors with its summation.

Accurate Concept Drift Detection Method (ACDDM) \cite{Yan2020ACDDM} uses Hoeffding’s inequality to analyze the inconsistency of the error rate for detecting concept drift. Lughofer \textit{et al.} \cite{Lughofer2016FPH} have designed an approach to detect concept drift in semi-supervised and fully unsupervised problems. The authors modified the standard Page-Hinkley test (PHT) \cite{Mouss2004PHT} to a faded version that outweighs older statistics. The PH statistic used to obtain classifier's confidence is based on the Hoeffding bound. Sakamoto \textit{et al.} \cite{Sakamoto2015PHT} have applied DDM to clustering problems by utilizing the assignment error and PHT was used to detect the changes. 

DDM was also integrated into more complex frameworks that cope with concept drift. A Meta-cognitive Recurrent Recursive Kernel Online Sequential Extreme Learning Machine with a modified DDM (meta-RRKOS-ELM-DDM) \cite{Liu2019Meta} was presented to solve the concept drift problem and reduce the learning time. The authors modified DDM so it could be employed in time series forecasting by calculating the error rate $ER_{l,p}$ and the standard deviation $SD_{l,p}$, for each sample $l$ in step $p$ of the time series prediction. The meta-cognitive learning strategy automatically finds the Approximate Linear Dependence Kernel Filter (ALD) threshold to scale down the computation complexity.

Follow-the-Regularized-Leader with Adaptive Decaying Proximal (FTRL-ADP) \cite{Huynh2018FTRL} is based on Time Decaying Adaptive Prediction (TDAP) algorithm and uses the DDM drift detector to speed up the adaptation to concept drift. This adaptation allows tuning the decaying rate of the TDAP algorithm, automatically. Online Map-Reduce Drift Detection Method (OMR-DDM) \cite{Andrzejak2012OMR} combines the online error rate of parallel classification algorithms to detect drifts using a Map-Reduce framework.

DDM was also modified to be utilized in online class imbalance learning problems. Drift Detection Method for Online Class Imbalance (DDM-OCI) \cite{Wang2013OCI} is one of the first algorithms in this category. The method uses the same test statistic as DDM, but tracks the degradation in the minority-class recall to signal concept drift. The method triggers many false alarms in scenarios where the majority-class is affected by the drift since it only considers the true positive rate $P(tpr)$. Linear Four Rates (LFR) \cite{Wang2015LFR} has improved the limitation of DDM-OCI by monitoring the four rates of the confusion matrix, true positive rate ($tpr$), true negative rate ($tnr$), false positive rate ($fpr$) and false negative rate ($fnr$). Hierarchical Linear Four Rates (HLFR) \cite{Yu2019HLFR} uses the same four rates as LFR hierarchically in two testing layers. PerfSim \cite{Antwi2012Perfsim} handles imbalanced datasets with concept drift by calculating the Cosine Similarity measure of $TP$ and $FP$ of all classes and comparing them to a given threshold $\alpha$.

Some other testing techniques were applied to monitor the model's performance degradation. Song \textit{et al.} \cite{Song2020FDA} have proposed fuzzy error deviation (fed) metric, which is computed to estimate the drift severity based on the variation of the predictor error. Adaptive Online Incremental Learning for evolving data streams (AOIL) \cite{Zhang2021AOIL} monitors the change in the mean and variance values of the loss error to detect the drift. Spectral Entropy Drift Detector (SEDD) \cite{Chikushi2021SEDD} computes the spectral entropy along the error stream to verify the fluctuation's magnitude along the learning process. The Drifter algorithm \cite{Oikarinen2021Drifter} calculates the generalization error on the dataset (RMSE) to detect concept drift. The algorithm determines the detection threshold $\sigma$ using receiver operating characteristics (ROC)
analysis.

EWMA for Concept Drift Detection (ECDD) \cite{Ross2012ECDD} adjusts the conventional exponentially weighted moving average charts (EWMA) \cite{Yeh2008EWMA} to monitor changes in the error rate of the classifier. At time $t$, the error rate $\hat{p}_{0,t}$, and the dynamic standard deviation $\sigma_{Z_{t}}$ of the EWMA estimator $Z_{T}$, are calculated.  Concept drift is flagged if:

\begin{equation}
   Z_{t} > \hat{p}_{0,t} + L_{t}\sigma_{Z_{t}}
    \label{eq5}
\end{equation}

%Bestoun --- 3 Sep 2021 ---- 6

where the control limit $L_{t}$ is provided by the authors. Disabato and Roveri \cite{Disabato2019ACNN} have adapted Convolutional Neural Networks (CNN) by incorporating Change Detection Tests (CDTs) based on monitoring the classification error to detect concept drift using CUmulative SUM (CUSUM) test \cite{Page1954CUSUM}. Other works control different performance metrics to detect drifts. As in \cite{Wang2020AUC}, authors have proposed a family of AUC-based metrics, namely Prequential Multi-Class AUC (PMAUC), Weighted AUC (WAUC), and Equal Weighted AUC (EWAUC). The metrics can be utilized as a part of the concept drift detection method for multi-class imbalanced data by tracking their values over time.

Extreme learning machine (ELM) \cite{Huang2006ELM} was exploited to detect concept drift, in particular, online sequential ELM (OS-ELM) \cite{Liang2006OSELM}. Yang \textit{et al.} \cite{Yang2019OSELMs} have proposed a method that can detect concept drift based on the dissimilarities between the output weights of the OS-ELM models for every chunk of new data. Dynamic Extreme Learning Machine (DELM) \cite{Xu2017DELM} modifies ELM by adding concept drift detection that monitors the performance degradation of the learner. Based on the result of the detector, DELM will add additional hidden layer nodes in case of concept drift occurrence. Another method that utilizes ELM is the Meta-cognitive online sequential extreme learning machine (MOS-ELM) \cite{Mirza2016MOS}. MOS-ELM incorporates two tests depending on the type of drift, one for gradual drift and another for sudden drifts. It uses the weighted extreme learning machine (WELM) to track the classification performance in imbalanced datasets.

%Bestoun --- 6 Sep 2021 ---- 7

\subsection{Windowing Technique}
Window-based detectors divide the data stream into windows based on data size or time interval in a sliding manner. These methods monitor the performance of the most recent observations introduced to the learner and compare it with the performance of a reference window.

ADaptive WINdowing (ADWIN) and its extension (ADWIN2) \cite{Bifet2007ADWIN} are among the most popular methods that use the windowing technique to detect drifts. ADWIN uses the Hoeffding bound to examine the change between the means, $\mu_{hist}$ and $\mu_{new}$, of the two \textit{sufficiently large} sub-windows, $W_{hist}$ and $W_{new}$:
\begin{equation}
    |\mu_{hist} - \mu_{new}| > 2\epsilon_{cut}
\end{equation}
where $\epsilon_{cut}$ is the optimal cut:
\begin{equation}
    \epsilon_{cut} = \sqrt{\frac{1}{2m}ln\frac{4|W|}{\delta}}
\end{equation}
where $m$ is the harmonic mean of the two windows, and $\delta$ is a pre-defined confidence parameter.

SEED \cite{Huang2014SEED} adopts the ADWIN method by comparing two sub-windows within a window $W$, a left sub-window $W_L$ and right sub-window $W_R$. SEED monitors a binary sequence of the classification decision, $1$ for correct predictions and $0$ for errors. The algorithm sets the boundaries of cutting the windows by using the Hoeffding Inequality with Bonferroni correction to calculate $\epsilon_{cut}$, the test statistic to compare the averages of data instances of each window. 

Another well-recognized and straightforward method for concept drift detection is STEPD \cite{Nishida2007STEPD}, which relies on two-time windows, a recent window $r$ and overall window $o$. It applies the statistical test of equal proportions to compare the accuracies between the two windows as follows:

\begin{equation}
T\left(r_{o}, r_{r}, n_{o}, n_{r}\right)=\frac{\left|r_{o} / n_{o}-r_{r} / n_{r}\right|-0.5\left(1 / n_{o}+1 / n_{r}\right)}{\sqrt{\hat{p}(1-\hat{p})\left(1 / n_{o}+1 / n_{r}\right)}}
\end{equation}

where $r$ is the number of correct predictions, $n$ is the window size, and $\hat{p} = \left(r_{o}+r_{r}\right) /\left(n_{o}+n_{r}\right)$. P-value is then calculated and compared with the significance level to signal the drift.
Wilcoxon Rank Sum Test Drift Detector (WSTD) \cite{De2018WSTD} was inspired by STEPD and applies Wilcoxon rank sum statistical test \cite{Wilcoxon1992Individual} to detect the drift and limits the size of the older window.
Cabral and Barros \cite{De2018FSDD} have modified STEPD to propose three methods to detect drifts, namely Fisher Proportions Drift Detector (FPDD), Fisher Square Drift Detector (FSDD), and Fisher Test Drift Detector (FTDD). The only difference between these methods and STEPD is that they used Fisher’s Exact test \cite{Fisher1922Test} to calculate the p-value. Cosine Similarity Drift Detector (CSDD) \cite{Hidalgo2019CSDD} works similarly to WSTD by calculating the confusion matrix based on the Positive Predictive Value (PPV) and False Discovery (FDR) rates instead of TP and FP for each window, which are calculated as $PPV_{r} = TP/(TP+FP)$ and $FDR_{r} = FP/(TP+FP)$. Then the Cosine Similarity is computed between the vectors created from the confusion matrices of the two windows to signal a drift or warning alert. In a recent study \cite{Wu2021Nacre}, authors have proposed the Nacre framework that uses the ADWIN strategy to set the window size in a stability detector that monitors the predictive performance.

Similar practices have been followed to process the window. McDiarmid Drift Detection Method (MDDM) \cite{Pesaranghader2018MDDM} slides a window over the prediction results, $1$ for correct predictions, and $0$ for false predictions. The entries of the prediction results stream are weighted by recency. The method uses McDiarmid’s inequality \cite{Mcdiarmid1989Method} to determine the significance in the difference between the maximum weighted average seen so far and the weighted mean of entries in the sliding window. ADaptive sliding window-based Detection Method (ADDM) \cite{Du2014ADDM} follows a similar approach as MDDM but monitors the entropy of the prediction results stream over a sliding window.

Other window-based approaches can be found in the literature. The Margin Density Drift Detection (MD3) \cite{Sethi2015PerfTr} approach processes the data stream as a sliding window. It monitors the number of samples that fall within the classifier's margins for every chunk of data. The approach triggers a drift alert based on a comparison with the density threshold $\theta$. Fast switch Naıve Bayes model (fsNB) \cite{Liu2020fsNB} performs two-sample Kolmogorov-Smirnov test (KS test) \cite{Kolmogorov1933Sulla} to compare the residuals of a fine-tuned model and a retrained model to decide which model to use. Error DISTance for drift detection and monitoring (EDIST) \cite{Khamassi2014EDIST} modifies EDDM by maintaining two data windows, a global sliding window and another one that contains the current examples. EDIST detects the drift by checking if the error distance distributions between the two windows exceed a threshold $\varepsilon$. The $\varepsilon$ value tunes itself adaptively based on the statistical hypothesis test. The experiments showed that the method is robust to noise and false alarms. Khamassi \textit{et al.} \cite{Khamassi2015EDIST2} have extended EDIST by introducing EDIST2, which can handle gradual local drifts by using all the data in relearning the model instead of only using the data window in the drifted region. Anti-concept Drift Detection Algorithm (ADDS) \cite{Liu2017ADDS} applies Hoeffding’s inequality to track the difference between the optimal accuracy and the real-time accuracy in a sliding window. ADDS concludes that the error in the classification accuracy should be within a threshold $\epsilon= \sqrt{\frac{1}{2n}ln\frac{1}{\delta}}$, where $n$ is the sliding window size and $\delta$ is the confidence level, otherwise concept drift is detected.

% andreas.kassler: should we say something about Hoeffdings inequality at the begin as this seem to be central to many approaches? Like It follows from Hoeffding's  inequality that the probability that the estimated and true values differ by more than t is bounded by e−2nt2.

%Bestoun --- 6 Sep 2021 ---- 8

\subsection{Ensemble Learning}

Concept drift detectors that are ensemble-based operate by combining the results of multiple diverse base learners. The overall performance is monitored by either considering the accuracy of all the ensemble members or the accuracy of each individual base learner. Note that this is different from an ensemble of drift detectors as in \cite{Maciel2015Lightweight, Du2015Selective}, where the decisions of multiple drift detectors are combined to signal the drift. Experimental studies demonstrate that an ensemble of drift detectors does not guarantee higher performance than the individual detection methods \cite{Wozniak2016Ensembles}.

Ensemble-based detectors trigger concept drift if the learners suffer from a significant level of performance degradation. This assumption is based on the fact that each learner has capabilities in solving specific problems \cite{Krawczyk2017Ensemble}. Most of the ensemble-based detectors are built upon the Weighted Majority Algorithm (WMA) \cite{Littlestone1994WMA} method. WMA elects the best learners in the ensemble by giving each one a weight based on its performance. Streaming Ensemble Algorithm (SEA) \cite{Street2001SEA} approach is one of the earliest ensemble-based works to tackle concept drift. SEA handles the drift implicitly by creating a new learner for each new chunk of the data till the maximum number of learners is reached. The learners are refined based on their prediction performance. A similar method in refining the ensemble was proposed in the Accuracy Weighted Ensemble (AWE) \cite{Wang2003AWE}. The novelty of AWE is in selecting the best learners by using a special version of the mean squared error that deals with probabilities to select the best $n$ learners and discard outdated learners with the highest performance degradation rate. Brzezinski and Stefanowski \cite{Brzezinski2011AUE1} have proposed Accuracy Updated Ensemble (AUE) algorithm, which improves AWE by conditionally updating the component learners rather than only regulating the weights. The authors also used a simpler weighting function than the one in AWE. AUE2 \cite{Brzezinski2013AUE2} improved AUE by introducing a cost-effective weight and pruning base learners. Online Accuracy Updated Ensemble (OAUE) \cite{Brzezinski2014OAUE} utilizes a drift detector included in an online learner that triggers a reweighting signal to the learner. Accuracy and Growth Rate updated Ensemble (AGE) \cite{Liao2014AGE} has extended AUE2 to react to various types of drift. AGE uses the geometric mean to design the Growth Rate of base learners.

Dynamic Weighted Majority (DWM) \cite{Kolter2007DWM} is one of the most popular passive ensemble approaches, which employs a weighting mechanism inspired by WMA. Every learner's weight is reduced by a multiplicative factor $\beta$, $0\leq\beta\leq1$, when it gives a wrong prediction every $\rho$ time step. To overcome the drawback of DWM, which does not consider the learner's performance on the training data, DWM-WIN was proposed in \cite{Mejri2013DWMWIN}. DWM-WIN is an ensemble method that includes the learner’s age in the weighting mechanism and tracks the concept drift in the learning phase. In recent research, Heterogeneous Dynamic Weighted Majority (HDWM) \cite{Idrees2020HDWM} was proposed to turn DWM into a heterogeneous ensemble by automatically choosing the best learners to be used over time to prevent performance degradation. Recurring Dynamic Weighted Majority (RDWM) \cite{Sidhu2019RDWM} is built upon DWM by forming two ensembles of learners. The primary ensemble represents the current concepts, and the secondary ensemble consists of the most accurate learners.

Another well-known ensemble-based drift detection method is Learn++.NSE (incremental learning for NSEs) \cite{Elwell2011NSE}. Learn++.NSE is the first version of the notable set of ensemble algorithms Learn++ \cite{Polikar2001Learn++} to address concept drift. In Learn++.NSE, a set of learners is trained on chunks of data examples. The training examples are weighted according to the ensemble error on this example. If the example is correctly classified by the ensemble $i$, Learn++.NSE sets its weight to 1, otherwise it is penalized to $w_i = 1/e$. The sigmoid function is used to weigh the learners in the ensemble based on their errors on the old and current chunks. Ditzler and Polikar \cite{Ditzler2012CDS} have proposed a framework that includes two related ensemble-based approaches, namely Learn++.CDS and Learn++.NIE. They extended their prior work on Learn++.NSE to accommodate class-imbalanced data. The methods monitor the performance of both the majority and minority classes. On-line Weighted Ensemble (OWE) \cite{Soares2015OWE} was proposed to adapt Learn++ for regression tasks. 

Other methods use the diversity between the learners in the ensemble. Diversity for Dealing with Drifts (DDD) \cite{Minku2011DDD} controls the diversity level of the learners in the ensemble by incorporating both low diversity and high diversity ensembles. The low diversity ensemble is used to detect the drift, and the high diversity ensemble is used after detecting the drift. Diversified Online Ensembles Detection (DOED) \cite{Sidhu2015DOED} develops two ensembles with different levels of diversity, \textbf{E0} and \textbf{E1}. DOED uses only one significance level to detect concept drift with \textbf{E0} and \textbf{E1} using the P-value. If any of the ensembles detect a drift, the ensemble is re-initialized. If both detect the drift, the ensemble with the lower accuracy is re-initialized. Recurrent Adaptive Classifier Ensemble (RACE) \cite{Museba2021RACE} preserves an archive of diverse learners and uses EDDM to detect recurring drifts. The online drift detector for the K-class problem (ODDK) \cite{Mahdi2021ODDK} was proposed to handle multi-class problems with concept drift. The algorithm constructs a contingency table that stores the variation of the diversity of a pair of classifiers and uses the PH test to detect concept drift.

The benchmark methods of the other categories, statistical process control and windowing technique, were also used in ensemble frameworks. Pinagé \textit{et al.} \cite{Pinage2020DCSLA} have modified DDM and EDDM to work as unsupervised detection methods producing a \textit{pseudo prequential error rate} that is monitored for every ensemble member by assuming the predicted value is the true label. The drift is detected if \textit{n} members of the ensemble reach a drift level. Predictive and parameter INsensitive Ensemble (PINE) \cite{Ang2012PINE} is an ensemble approach that processes asynchronous concept drifts in classification in distributed networks. A modified version of the ADWIN drift detector is provided for each peer of the framework. The detector monitors a stream of accuracies represented by ones and zeros. More recently, Liu\textit{ et al.} \cite{Liu2021CALMID} have proposed CALMID method for multiclass imbalanced streaming data with concept drift that uses ADWIN algorithm in ensemble settings. Associative Classification over Concept Drifting Data Streams (ACCD) \cite{Waiyamai2014ACCD} checks the current accuracy of an ensemble of online classifiers by comparing it with the estimated statistical lower bound of maximum accuracy to signal a drift. EnsembleEDIST2 \cite{Khamassi2019Ensemble} makes use of EDIST2 as a drift detector in the proposed ensemble-based drift handling approach to track the learners' performance.

% AK Sep 4

%Bestoun --- 6 Sep 2021 ---- 9

Predict-Detect streaming framework \cite{Sethi2018Handling} relies on detecting adversarial drifts from unlabeled data streams inspired by the MD3 framework. The framework uses the training data to learn the expected disagreement $PD_{Ref}$ and accepted deviation $\sigma_{Ref}$ of the ensemble. An adversarial drift is detected if a sudden increase in the disagreement metric \textit{PD} occurs. Efficient Concept Drift and Concept Evolution Handling over Stream Data (ECHO) \cite{Haque2016ECHO} is a semi-supervised ensemble-based framework that contains a concept drift detection technique. ECHO maintains a sliding window over the data stream to monitor significant changes in the classifier's confidence to detect concept drift using the CUSUM test. Khezri \textit{et al.} \cite{Khezri2021SSE} proposed an ensemble-based Performance-Based Selection (PBS) metric for semi-supervised learning problems with concept drift. The model performance is evaluated based on pseudo-accuracy and energy regularization.

ELM has also been employed in the ensemble approach to deal with concept drift. An ensemble of online sequential extreme learning machines (ESOS-ELM) \cite{Mirza2015ESOS} was proposed to tackle concept drift in class imbalance data. ESOS-ELM maintains an ensemble of OS-ELMs and monitors the error rate using a threshold-based technique. In\cite{Oliveira2017PSO}, authors have developed two approaches IDPSO-ELM-B and IDPSO-ELM-S to detect concept drift in time series forecasting. The approaches were built upon the swarm behavior of ELM by using the ECDD approach. Xu \textit{et al.} \cite{Xu2017ALELM} have proposed an alternating learners framework that uses a drift detector and employs ELM as a base learner for regression problems.

Other strategies were proposed in an ensemble learning framework to deal with concept drift. Number and Distance of Errors (NDE) \cite{Dehghan2016NDE} is an ensemble method that detects concept drift based on the number and distance between the errors and compares it with a threshold. Knowledge-maximized ensemble (KME) \cite{Ren2018KME} is a concept-drift-detection system that contains a $TEST_l$ concept drift detector which checks if the classification error of the ensemble falls below the confidence interval in a sliding window. Enhanced Concept Profiling Framework (ECPF) \cite{Anderson2019ECPF} is a meta-learning framework that tracks the learner behavior to detect changes. Wang \textit{et al.} have proposed a new pruning criterion, called the loss improvement ratio (LIR), for performance evaluation that is utilized in a pruning strategy to remove outdated learners. The meta-learner decides if the current learner should be reused or replaced based on the performance. Weighted classification and Update algorithm of Data stream based on Concept Drift Detection (WUDCDD) \cite{Zhang2019WUDCDD} approach signals a drift warning if the performance degrades for the current data chunks. The system signals a drift detection alert if the degradation is still present. The method calculates the Mahalanobis distance between the classification error rate on the data blocks. \cite{Yazdi2020UECM} uses the Uncertainty Error Correlation Matrix (UECM) to detect concept drift and give each online learner a corresponding weight. UECM is constructed from the error value of the online learning algorithms in the ensemble, and each entry of the matrix represents the strength between the loss function of each learner. 

%Bestoun --- 6 Sep 2021 ---- 10

\section{Analysis and Discussion}\label{analysis}
This section analyzes and discusses the relevant works we have reviewed to draw conclusions and accentuate the trends in this research area. The analysis is based on spotlighting the main facets that characterize the methods presented in this paper. To facilitate answering \textbf{RQ3}, we have steered our attention to investigate multiple attributes that were handled in designing the methods. The attributes are: (1) the machine learning problem managed by the method, (2) the performance metric used to track model degradation, (3) the base learner employed in the predictive system, and (4) the type of drift addressed by the approach.

\subsection{Machine Learning Problem Scope}
\begin{figure}
\centering\includegraphics[width=0.85\linewidth]{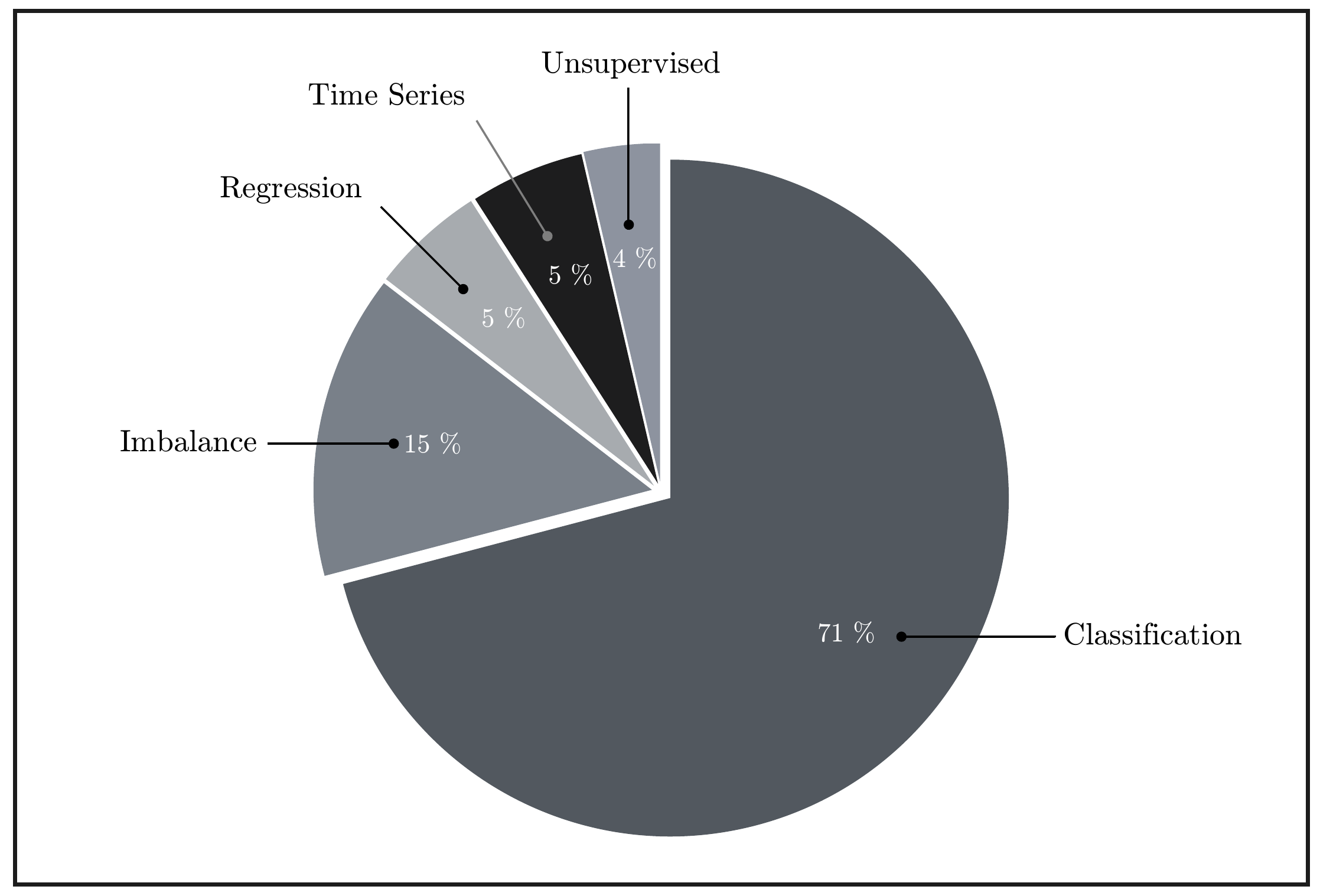}
\caption{Machine learning problem scope of the methods}
\label{ML_problems}
\end{figure}

Narrowing down the scope of the machine learning problem is a fundamental step in designing the concept drift detection method since each learning problem requires calculating different performance metrics. In Figure \ref{ML_problems} we summarized the machine learning scope of the surveyed methods. We can see that drift detection for classification tasks is the main scope in the literature, while few approaches address the regression settings. The main reason for that is the lack of relevant datasets for regression problems with concept drift \cite{Cavalcante2016FEDD}, and the wide availability of classification datasets for concept drift detection purposes \cite{Lu2016LearningUnder}. There is also a reasonable number of studies in the area of class imbalance problems since the concept drift, and class imbalance problems are closely related and affect each other \cite{Wang2018Imbalance}. More recently, performance-based drift detection methods have been proposed in the context of more complicated, semi-supervised, and unsupervised problems. These problems pose a significant challenge for performance-based detectors since the ground truth labels are not provided. Semi-supervised detectors usually operate by predicting the labels of the unlabeled examples and proceed with computing the performance loss to detect drift \cite{Ditzler2011Semi}. On the contrary, the unsupervised drift detector approach tries to estimate a pseudo-error and self-evaluate the performance \cite{Cerquitelli2019Towards}.

Most of the proposed approaches are devoted to solving a specific problem. This confirms that concept drift detection adheres to the No Free Lunch Theorem \cite{Ho2002NFL}, and a universal approach that copes with all machine learning problems are challenging to find \cite{Hu2020NoFLT}.

\subsection{Performance Metrics}
\begin{figure}
\centering\includegraphics[width=0.9\linewidth]{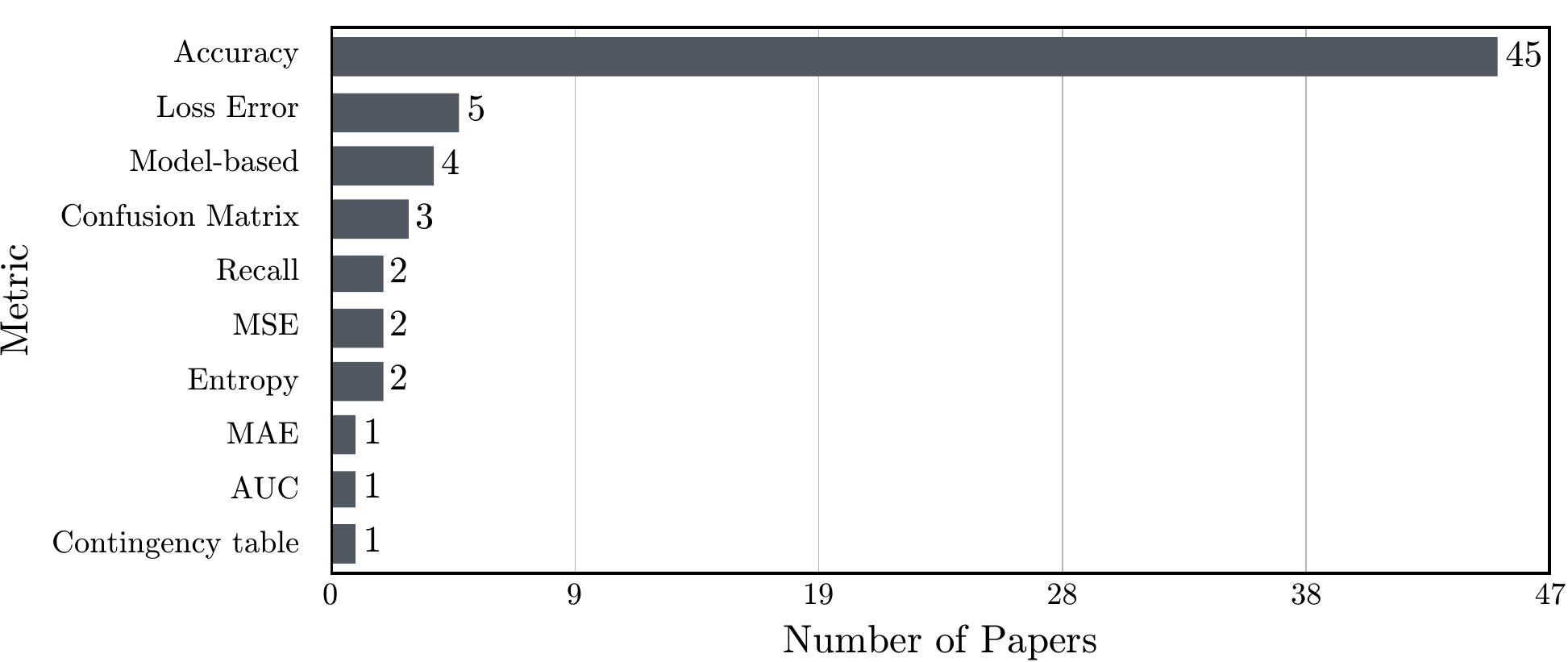}
\caption{Performance metrics monitored in the methods}
\label{Metrics}
\end{figure}

As shown in Figure \ref{Metrics}, the majority of the methods rely on the classification error rate to detect the degradation in the predictive performance. This could be because most approaches have been evaluated within the classification task context, which received most of the attention in the literature. In addition, calculation of classification error rate metric entails low complexity and cost needed. Since the accuracy is not always indicative of performance loss, drift detectors are criticized for a high number of false alarms. For class imbalance tasks, and since accuracy is not an expressive metric for performance, authors have adopted other metrics such as confusion matrix and AUC. Some studies have designed drift detectors based on metrics calculated from the model's intrinsic behavior, such as performance gain and growth rate. We have grouped these model-wise metrics into a model-based category.

\subsection{Base Learners}
\begin{figure}
\centering\includegraphics[width=1.0\linewidth]{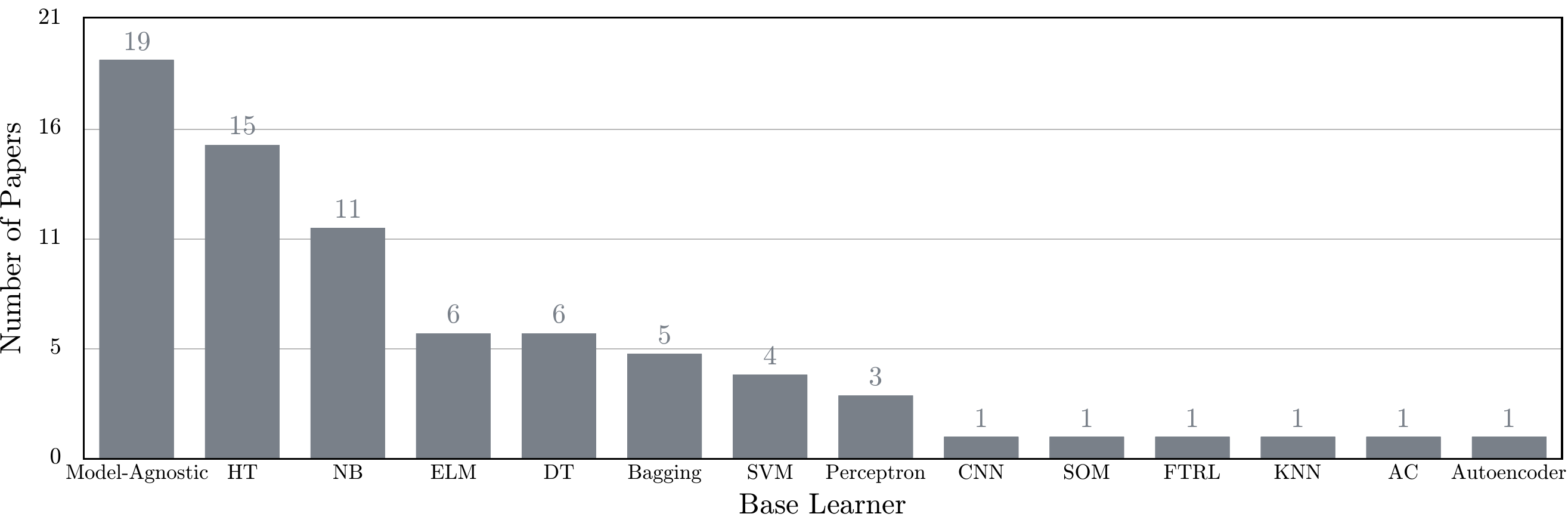}
\caption{Base learners adopted in the methods}
\label{learner}
\end{figure}

Drift detection systems require many updates once they are deployed. Consequently, drift detection methods should support incremental learning and adapt dynamically. For that reason, Hoeffding Trees (HT) and Naive Bayes (NB) are adopted as base learners for the majority of performance-based concept drift detectors. Furthermore, HT and NB also have sufficient capability to learn from and deal with massive data streams. More recent works have adopted neural networks within their framework. Still, these approaches could make deploying within big data stream systems challenging since it is difficult to update the neural network architecture dynamically. Another major drawback for neural networks is the lack of transparency and interpretability \cite{Buhrmester2019Analysis, Wang2019Neural}. This drawback causes a burden to concept drift handling systems since drift understanding plays a significant part in detecting and adapting to drifts \cite{Lu2016LearningUnder, Lu2020Data}. This Figure \ref{learner} summarizes the base learners used in the reviewed approaches. We use \textit{model-agnostic} only for those papers which explicitly stated that the detection method could be integrated with any predictive model. Otherwise, we use the base learner as reported in the original study. 

\subsection{Drift Types}
Table \ref{paper_dtypes} provides an overview of the methods that explicitly mentioned the handled drift type. Sudden and gradual drifts are the main drift types addressed in the reviewed literature. While fewer works addressed the incremental drift, it is not always easy to distinguish between the natural evolution of systems and continuous changes. Recurring drift must be processed in a specific way, where the system must be supplied with a buffer to store the old behavior and reuse the learned knowledge from the past observations once it reappears in the system.

\afterpage{
\begin{table}
\thispagestyle{empty}
\begin{center}
\footnotesize
\renewcommand{\arraystretch}{1}
\setcounter{table}{3} 
\caption{Summary of the methods with the handled drift type. The method names in italics were proposed by authors since there was no name given in the original paper.}
\label{paper_dtypes}
\begin{tabular}{|c|c|c|c|c|c|}
\hline 
\multirow{2}{*}{Method} & \multirow{2}{*}{Year} & \multicolumn{4}{c|}{Drift Type}\tabularnewline
\cline{3-6} \cline{4-6} \cline{5-6} \cline{6-6} 
 &  & Sudden & Gradual & Incremental & Recurring\tabularnewline
\hline 
SSE-PBS \cite{Khezri2021SSE} & 2021 & \checkmark &  \checkmark &  &   \tabularnewline
\hline 
ODKK \cite{Mahdi2021ODDK} & 2021 &\checkmark  & \checkmark  &  &  \checkmark \tabularnewline
\hline 
RACE \cite{Museba2021RACE} & 2021 &  &   &  &  \checkmark \tabularnewline
\hline 
LIR-eGB \cite{Wang2021LIR} & 2021 &  \checkmark&   \checkmark&  \checkmark&   \tabularnewline
\hline 
CALMID \cite{Liu2021CALMID} & 2021 & \checkmark &  \checkmark & \checkmark &  \checkmark \tabularnewline
\hline 
Nacre \cite{Wu2021Nacre} & 2021 &  &   &  &  \checkmark \tabularnewline
\hline 
SEDD \cite{Chikushi2021SEDD} & 2021 & \checkmark &  \checkmark &  &  \tabularnewline
\hline 
\textit{OFE-UECM} \cite{Yazdi2020UECM} & 2020 & \checkmark & \checkmark & \checkmark & \checkmark\tabularnewline
\hline
FDA \cite{Song2020FDA} & 2020 & \checkmark &   & \checkmark &  \tabularnewline
\hline 
ACDDM \cite{Yan2020ACDDM}& 2020 & \checkmark & \checkmark & \checkmark & \checkmark\tabularnewline
\hline 
HDWM \cite{Idrees2020HDWM} & 2020 & \checkmark & \checkmark &  & \checkmark\tabularnewline
\hline 
\textit{OS-ELMs} \cite{Yang2019OSELMs} & 2020 & \checkmark & \checkmark & \checkmark & \tabularnewline
\hline 
\textit{DCS-LA} \cite{Pinage2020DCSLA} & 2020 & \checkmark & \checkmark &  & \tabularnewline
\hline 
HLFR \cite{Yu2019HLFR} & 2019 & \checkmark & \checkmark & \checkmark & \checkmark\tabularnewline
\hline 
RDWM \cite{Sidhu2019RDWM} & 2019 & \checkmark & \checkmark & \checkmark & \checkmark\tabularnewline
\hline 
CSDD \cite{Hidalgo2019CSDD} & 2019 & \checkmark & \checkmark &  & \tabularnewline
\hline 
ECPF \cite{Anderson2019ECPF} & 2019 &  &  &  & \checkmark\tabularnewline
\hline 
FHDDMS,FHDDMS\textsubscript{add} \cite{Pesaranghader2018FHDDMS} & 2018 & \checkmark & \checkmark &  & \tabularnewline
\hline 
FPDD, FSDD \cite{De2018FSDD} & 2018 & \checkmark & \checkmark &  & \tabularnewline
\hline 
FTRL-ADP \cite{Huynh2018FTRL}  & 2018 & \checkmark & \checkmark & \checkmark & \checkmark\tabularnewline
\hline 
KME-TEST\textsubscript{l} \cite{Ren2018KME} & 2018 & \checkmark & \checkmark & \checkmark & \checkmark\tabularnewline
\hline 
WSTD \cite{De2018WSTD} & 2018 & \checkmark & \checkmark &  & \tabularnewline
\hline 
MDDM \cite{Pesaranghader2018MDDM} & 2018 & \checkmark & \checkmark &  & \tabularnewline
\hline 
RDDM \cite{Barros2017RDDM} & 2017 & \checkmark & \checkmark &  & \tabularnewline
\hline 
ADDS \cite{Liu2017ADDS} & 2017 & \checkmark & \checkmark &  & \tabularnewline
\hline 
\textit{AL-ELM} \cite{Xu2017ALELM} & 2017 & \checkmark & \checkmark &  & \tabularnewline
\hline 
\textit{FPH-DD} \cite{Lughofer2016FPH} & 2016 & \checkmark & \checkmark &  & \tabularnewline
\hline 
MOS-ELM \cite{Mirza2016MOS} & 2016 & \checkmark & \checkmark &  & \tabularnewline
\hline 
NDE \cite{Dehghan2016NDE} & 2016 & \checkmark & \checkmark &  & \tabularnewline
\hline 
FHDDM \cite{Pesaranghader2016FHDDM} & 2016 & \checkmark & \checkmark &  & \tabularnewline
\hline 
DOED \cite{Sidhu2015DOED} & 2015 & \checkmark & \checkmark & \checkmark & \checkmark\tabularnewline
\hline 
HDDM \cite{Frias2014HDDM} & 2015 & \checkmark & \checkmark &  & \tabularnewline
\hline 
ESOS-ELM \cite{Mirza2015ESOS} & 2015 & \checkmark & \checkmark &  & \tabularnewline
\hline 
LFR \cite{Wang2015LFR} & 2015 & \checkmark & \checkmark &  & \tabularnewline
\hline 
\textit{DDM-PHT} \cite{Sakamoto2015PHT} & 2015 & \checkmark & \checkmark & \checkmark & \tabularnewline
\hline 
EDIST \cite{Khamassi2014EDIST} & 2014 & \checkmark & \checkmark &  & \tabularnewline
\hline 
OAUE \cite{Brzezinski2014OAUE} & 2014 & \checkmark & \checkmark & \checkmark & \checkmark\tabularnewline
\hline 
SEED \cite{Huang2014SEED} & 2014 & \checkmark & \checkmark &  & \tabularnewline
\hline 
AGE \cite{Liao2014AGE} & 2014 & \checkmark & \checkmark &  & \tabularnewline
\hline 
ACCD \cite{Waiyamai2014ACCD} & 2014 & \checkmark & \checkmark &  & \tabularnewline
\hline 
ADDM \cite{Du2014ADDM} & 2014 & \checkmark & \checkmark &  & \tabularnewline
\hline 
AUE2 \cite{Brzezinski2014OAUE} & 2014 & \checkmark & \checkmark &  & \tabularnewline
\hline 
LEARN++.CDS \cite{Ditzler2012CDS} & 2013 & \checkmark & \checkmark & \checkmark & \checkmark\tabularnewline
\hline 
ECDD \cite{Ross2012ECDD} & 2012 & \checkmark & \checkmark &  & \tabularnewline
\hline 
\end{tabular}
\end{center}
\end{table}
\clearpage
}

\section{Conclusions and Future Directions}\label{conclusions}
Concept drift and performance degradation are two intertwined phenomena in predictive systems. The existence of one phenomenon articulates the other. In this paper, we presented a comprehensive and up-to-date overview of the concept drift research field. We started by describing the main causes of concept drift, followed by common definitions and measures of concept drift. We have compiled the various terms used in the literature to refer to concept drift types since the area is deluged with terminologies. We then presented concept drift detection approaches that track the performance degradation to identify changes. These performance-based methods work reversely by signaling concept drift when the performance degrades to a certain threshold. Real concept drift leads to deterioration in the predictive accuracy as it requires adaptation to changes. The findings of this study are extracted and summarized in the following points:
\begin{enumerate}
    \item Multiple terms can be found in the literature for the same concept drift type. Also, the same term is used for multiple concept drift types. Therefore we suggest using the mathematical definition to refer to specific concept drift types.
    \item The classification problem comprises the major part of the task scope in drift handling. A limited number of works have been developed to undertake other scopes. 
    \item Most existing performance-based detectors rely on monitoring the error rate to identify the performance degradation and trigger a drift; recent advances have monitored new performance metrics.
    \item Performance-based detection methods have been used in unsupervised and semi-supervised learning by introducing new metrics to evaluate the model's performance, such as \textit{pseudo-error}.
    \item Most of the designed solutions used Hoeffding Trees or Naive Bayes algorithm as base learners. While employing neural networks has recently started to emerge.
    \item There is still no clear evidence about the ideal drift detector to be used in a specific problem or setting.
\end{enumerate}

Based on the mentioned findings, we suggest the following future research directions:

\begin{enumerate}
    \item Since few methods deal with regression settings, more research on detecting drifts in regression scope is highly desired. It is considered one of the main tasks in machine learning and is now employed in a wide range of applications \cite{Sarker2021Machine}.
    \item As previously proven in the comparison studies \cite{Barros2019Ens, Barros2018Large}, there is no single drift detector that works better than all the others in all scenarios. It would be interesting to evaluate the methods against different datasets and investigate their applicability in specific domains. This would support users in selecting the suitable method for the problem at hand.
    \item Most of the existing methods in the literature suffer from a high number of false alarms. This is because most of the approaches are over-reliant on monitoring the degradation in the learner's accuracy. A multiple hypothesis technique could be a solution by monitoring other metrics to have a stronger assumption on drift detection.
    \item The extensive research conducted on incremental and online learning paradigms could be leveraged in drift detection methods by employing the recent advances in drift handling systems \cite{Lobo2020Spiking}. Since these paradigms are characterized by high capabilities in continuously adapting to accommodate the incoming data points \cite{Cao2021Knowledge}.
    \item Another opportunity would be utilizing the staggering progress in the explainable deep learning field that has been recently achieved \cite{Bai2021Explainable, Chen2020Concept}. These explainable models would make efficient deep learning more useful in understanding and handling concept drift.
    
\end{enumerate}

% Firas -- DONE

\section*{Acknowledgement}

Parts of this work has been funded by the Knowledge Foundation of Sweden (KKS) through the Synergy Project AIDA - A Holistic AI-driven Networking and Processing Framework for Industrial IoT (Rek:20200067).

\clearpage
\scriptsize
\bibliographystyle{elsarticle-num}
\bibliography{bibliography.bib}
\end{document}